\documentclass[sigconf]{acmart}
\usepackage{amsmath}
\usepackage{colortbl}
\usepackage{xcolor}
\usepackage{makecell}
\usepackage{soul}

\AtBeginDocument{%
  \providecommand\BibTeX{{%
    \normalfont B\kern-0.5em{\scshape i\kern-0.25em b}\kern-0.8em\TeX}}}

\begin{document}

\title{PoKE: Prior Knowledge Enhanced Emotional Support Conversation with Latent Variable}

% \author{Xiaohan Xu}
% \email{xuxiaohan@iie.ac.cn}
% \affiliation{%
%   % \streetaddress{P.O. Box 1212}
%   % \city{Dublin}
%   % \state{Ohio}
%   % \country{USA}
%   % \postcode{43017-6221}

\author{Xiaohan Xu}
\affiliation{%
  \institution{Institute of Information Engineering \\ Chinese Academy of Sciences}
  \city{Beijing}
  % \state{Zhejiang}
  \country{China}
}
\email{xuxiaohan@iie.ac.cn}

\author{Xuying Meng}
\affiliation{%
  \institution{Institute of Computing Technology \\ Chinese Academy of Sciences}
  \city{Beijing}
  % \state{Zhejiang}
  \country{China}
}
\email{mengxuying@ict.ac.cn}

\author{Yequan Wang}
\authornote{Corresponding author.}
\affiliation{%
  \institution{Beijing Academy of Artificial Intelligence}
  \city{Beijing}
  % \state{Zhejiang}
  \country{China}
}
\email{tshwangyequan@gmail.com}

\begin{abstract}

Emotional support conversation (ESC) task can utilize various \textit{support strategies} to help people relieve emotional distress and overcome the problem they face, which has attracted much attention in these years. 
The emotional support is a critical communication skill that should be trained into dialogue systems.
Most existing studies predict the \textit{support strategy} according to current context to guide response.
However, most state-of-the-art works rely heavily on external commonsense knowledge to infer the mental state of the user in every dialogue round. 

Although effective, they may suffer from significant human effort, knowledge update and domain change in a long run.

Therefore, in this article, we focus on exploring the task itself without using any external knowledge. 
We find all existing works ignore two significant characteristics of ESC.
(a) Abundant \textit{prior knowledge} exists in historical conversations, such as the responses to similar cases and the general order of support strategies, which has a great reference value for current conversation.
% guiding supporter to explore user's problem and decide the target support strategy, 
(b) There is a \textit{one-to-many mapping relationship} between context and support strategy, i.e.multiple strategies are reasonable for a single context. 
It lays a better foundation for the diversity of generations.
Taking into account these two key factors, we propose \textbf{P}ri\textbf{o}r \textbf{K}nowledge \textbf{E}nhanced emotional support model with latent variable, \textbf{PoKE}.
versations as exemplars to guide generation, applies the first-order Markov model to help predict the target strategy, and then utilizes a latent variable to model the one-to-many relationship of support strategy. 
The proposed model fully taps the potential of prior knowledge in terms of exemplars and strategy sequence instead of external knowledge, and then utilizes a latent variable to model the one-to-many relationship of strategy.
Furthermore, we introduce a memory schema to incorporate the encoded knowledge into decoder. 
Experiment results on benchmark dataset show that our PoKE outperforms existing baselines on both automatic evaluation and human evaluation. Compared with the model using external knowledge, PoKE still can make a slight improvement in some metrics. Further experiments prove that abundant prior knowledge is conducive to high-quality emotional support, and a well-learned latent variable is critical to the diversity of generations.

\end{abstract}



\maketitle
\begin{figure}[t!] 
  \centering
  \includegraphics[width=\linewidth]{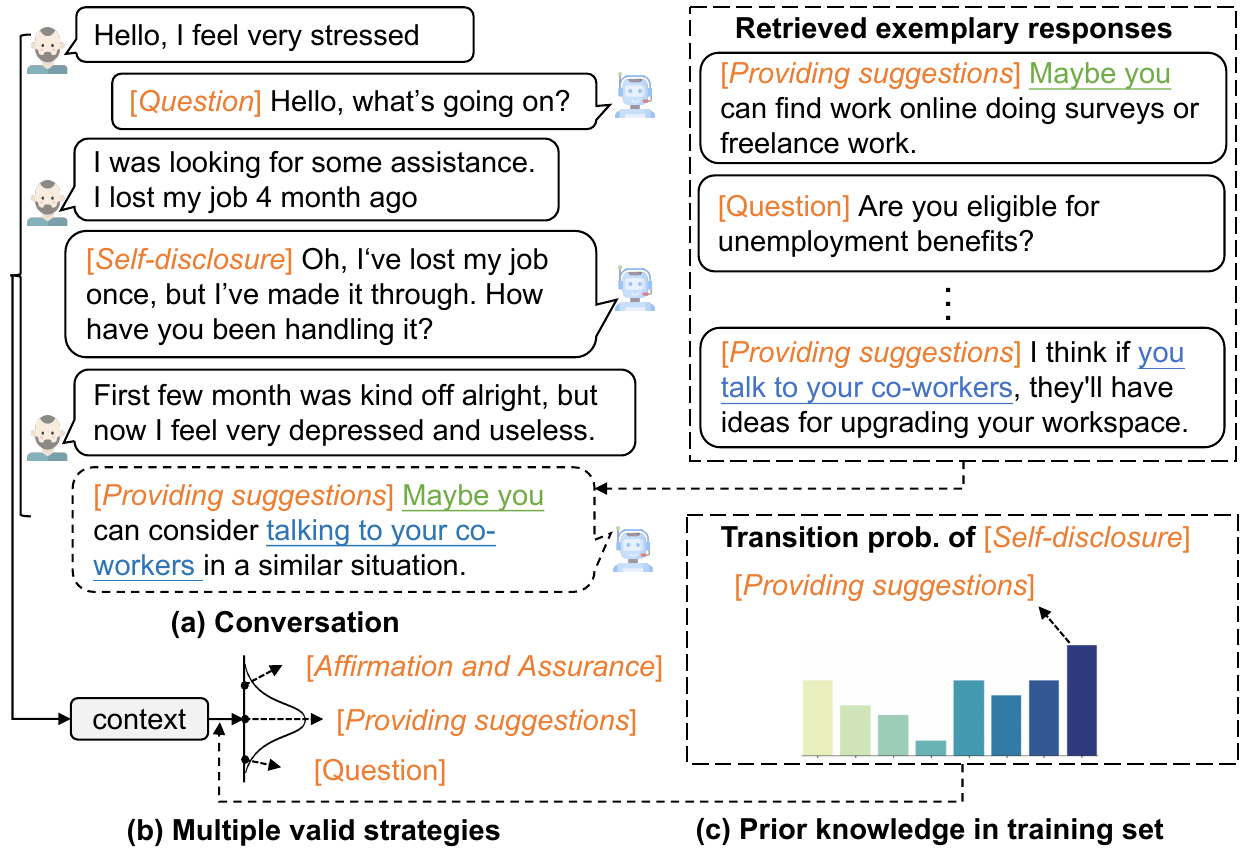}
  \caption{(a) An example to illustrate the ESC task. 
  (b) The one-to-many mapping relationship that there exist multiple valid strategies for a single context.
  (c) Retrieved exemplary responses give supporter more clues to focus on seeker's problem and express strategy more accurately. Meanwhile, transition probability of strategy provides a good bias to take a correct strategy.  \textcolor{orange}{Orange text} denotes the strategy taken by supporter.} 
  \label{fig:intro}
  % \vspace{-5mm}
\end{figure}
\section{Introduction}
Emotional support conversation (ESC)  \cite{DBLP:conf/acl/LiuZDSLYJH20} is an emerging and challenging task that devotes to coping effectively with help-seeker's emotional distress and helping them overcome the challenges they face. 
In general, a well-designed ESC system is crucial for many applications, e.g. customer service chats, mental health support, etc.~ \cite{DBLP:conf/acl/LiuZDSLYJH20}.
Compared to the well-researched emotional and empathetic conversation  \cite{DBLP:conf/emnlp/LinMSXF19, DBLP:conf/emnlp/MajumderHPLGGMP20, DBLP:conf/aaai/SabourZH22},
ESC focuses on reducing users' emotional stress using various \textit{emotional support strategies}, such as \textit{Question, Providing Suggestions}, etc.

Recently, several works have been proposed to explore the ESC task. 
BlenderBot-Joint  \cite{DBLP:conf/acl/LiuZDSLYJH20} generates a strategy token as a prompt to 
guide the desired response. 
MISC \cite{DBLP:conf/acl/TuLC0W022} uses an off-the-shelf generative commonsense model, called COMET \cite{bosselut-etal-2019-comet}, to infer the user's mental status, where the COMET can be seen as an external commonsense knowledge base. Then, MISC encodes them additionally and fuses multiple strategies into one response to generate skillfully.
GLHG \cite{DBLP:conf/ijcai/00080XXSL22} also utilizes COMET to generate the local intention of seeker in each dialogue round, but considers the hierarchical relationship between the seeker's global situation (summarizing the condition of the seeker) and the local intention.
Although effective, the commonsense knowledge in COMET need to be carefully integrated into these models to realize their best potential, and the external knowledge base requires a great deal of effort to develop. 
Further, their model may not be applicable when knowledge base is updated or application domain is changed. 
Therefore, in this article, we emphasize on exploring the existing knowledge in the dataset and the characteristics of ESC task under the setting of no external knowledge. 

Due to the characteristics of ESC, all existing works still suffer two key issues.
% However, 
First, 
all of them are limited to the scope of the current conversation, but ignore the abundant prior knowledge in global historical conversations. 
% Moreover, they ignore a special characteristic of ESC, i.e. one-to-many mapping relationship of support strategy. 
Moreover, they fail to model the one-to-many mapping relationship of strategy, i.e. not only one but multiple strategies could be valid for a single context. 
These issues lead to the challenge of generating high-quality and diverse responses. We next explain these two issues separately.

Generally, when we attempt to solve help-seeker's problems, we are adept in drawing on related prior knowledge as reference, e.g. psychologists would consult many prior classical cases relevant to current case  \cite{mieg2001social}. 
In ESC, instead of external knowledge, there also exists much prior knowledge to rely on, such as the (1) exemplary responses to similar cases and (2) the general order of support strategies. This prior knowledge has a great reference value to help explore seeker's problem and decide the target support strategy. 
An explanatory example in Figure \ref{fig:intro} illustrates how prior knowledge guides and benefits emotional support conversation. 
(1) The retrieved context-related responses from historical conversations, called \textit{exemplars}, can serve as prior knowledge of response.
On the one hand, some exemplars, e.g. ``I think if you talk to ...'', guide supporter to give more emphasis on the key problem ``losing job'', and thus benefit supporter to focus on and explore seeker's problem. On the other hand, some exemplars, e.g. ``Maybe you can find ...'', provide a hint to accurately express the target strategy \textit{Providing suggestions} in the sentence pattern starting with ``Maybe you''. 
(2) In addition to prior knowledge of response, the transition probability of strategy calculated in training set can act as prior knowledge to help decide the current strategy. This is because the support strategies in ESC follow the procedure of three stages (\textit{Exploration}, \textit{Comforting} and \textit{Action})  \cite{hill2009helping}. 
Figure \ref{fig:intro}(c) shows a transition probability of strategy \textit{Self-disclosure}. It illustrates that after sharing the similar difficulties they faced, supporters tend to use \textit{Providing suggestions} to give advice based on their experience.

Additionally, it is well known that dialogue systems have a one-to-many problem of generation, i.e. given a single context there exists multiple valid responses  \cite{DBLP:conf/acl/ZhaoZE17}.
In ESC, the supporter is required to take reasonable strategies, so there is also a one-to-many problem of support strategy. 
As shown in Figure \ref{fig:intro} (b), after the seeker states his problem, the supporter can also employ other valid strategies except for the frequently used strategy \textit{Providing suggestions}. Taking the strategy \textit{Question} to take a deeper look at user's problem or \textit{Affirmation and Reassurance} to comfort the user is also a decent choice. Moreover, adopting various strategies is beneficial to diverse responses.
In a nutshell, incorporating prior knowledge and modeling the one-to-many mapping relationship of strategy are critical to provide emotional support in ESC task.

To take into account these two significant characteristics of ESC, we propose a novel model called \textbf{P}ri\textbf{o}r \textbf{K}nowledge \textbf{E}nhanced emotional support conversation with latent variable model~(\textbf{PoKE}). The proposed model could not only fully tap the potential of prior knowledge in terms of exemplars and strategy sequence, but also model the one-to-many mapping relationship of strategy. 
First, we construct prior knowledge of exemplars and strategy sequence before training. Then we use a fine-tuned dense passage retrieval (DPR)  \cite{karpukhin2020dense} to retrieve a set of responses semantically related to the input context, and build a first-order Markov transition matrix of strategy sequence from training set. 
To model the one-to-many mapping relationship of strategy, we introduce conditional variational autoencoder (CVAE)  \cite{sohn2015learning} to predict diverse probability distribution of strategy conditioned on current conversation and prior knowledge of strategy sequence. Furthermore, we assign exemplars with different attentions according to the distribution of strategy to emphasize those more relevant exemplars. Lastly, we apply the technique of memory schema to effectively incorporate encoded prior knowledge and latent variable into decoder for generation.

\textbf{The key contributions are summarized as follows:}
% (1) We explore the emotional support conversation task without using any external knowledge base and propose a novel moedel, \textbf{PoKE}. PoKE can promote emotional support conversation by effectively modeling the prior knowledge in terms of exemplars and strategy sequence, and the one-to-many mapping relationship of strategy. 
(1) We explore the emotional support conversation task under the setting of no external knowledge base and propose a novel model, \textbf{PoKE}. PoKE can promote emotional support conversation by effectively modeling the prior knowledge in terms of exemplars and strategy sequence, and the one-to-many mapping relationship of strategy. 
% (2) We utilize prior knowledge to facilitate exploring user's problem and making decision of strategy, and utilize strategy distribution to pay more attention on strategy-related exemplars at sequence-level to improve expression of strategy.
(2) We utilize strategy distribution to denoise the exemplars and apply a memory schema to effectively incorporate encoded information into decoder.
(3) Experiments on benchmark dataset~(i.e., ESConv) of ESC task demonstrate that our method is superior to existing baselines on both automatic evaluation and human evaluation. Compared with the model using external knowledge, PoKE still can make a slight improvement in some metrics.
(4) Importantly, we reveal that abundant prior knowledge is conducive to high-quality emotional support, and a well-learned latent variable is critical to the diversity of generations.

\section{Related Work}
% In this section, we introduce present works related to our proposed PoKE.
In this section, we first detail some existing proposed methods for the emotional support conversation.
Then, because we utilize retrieved exemplars to guide generation and take a latent variable to solve the one-to-many issue of strategy, we will elaborate retrieve-based generation and one-to-many issue in dialogue system.
% te this common issue in dialogue system.
% exemplars are retrieved as prior knowledge to guide generation, which is a technique of retrieve-based generation. Besides, PoKE takes a latent variable to solve the one-to-many issue of strategy, so we elaborate this common issue in dialogue system.
\subsection{Emotional Support Conversation}
Before the task ESC is proposed, there are two relevant well researched dialogue systems, i.e. emotional chatting  \cite{DBLP:conf/aaai/ZhouHZZL18,Wei2019,song-etal-2019-generating} and empathetic responding  \cite{DBLP:conf/acl/RashkinSLB19, DBLP:conf/aaai/LinXWSLSF20, DBLP:conf/emnlp/LinMSXF19, DBLP:conf/emnlp/MajumderHPLGGMP20}. 
Emotional chatting needs to respond in appropriate emotion or the given emotion, such as happy or angry  \cite{DBLP:conf/aaai/ZhouHZZL18}. 
Empathetic responding needs to understand and feel what user is experiencing, and respond with empathy  \cite{DBLP:conf/acl/RashkinSLB19}. 
Compared with them, the emerging task of ESC aims at reducing help-seeker's emotional stress and help them explore and overcome the problem the face. 
% During the process of ESC, the supporter needs to identify seeker's problem, comfort the seeker, and provide some suggestions, and thus they have to adopt various support strategies with conversation goes  \cite{DBLP:conf/acl/LiuZDSLYJH20}. 
The first work on ESC task, called BlenderBot-Joint, adopts a chitchat bot BlenderBot \cite{roller2021recipes} as backbone and takes emotional support into account in conversation \cite{DBLP:conf/acl/LiuZDSLYJH20}. Specifically, they encode the context history and predict a strategy token. Then, they concatenate the predicted strategy token to the head of generation to guide the desired response.
%  \cite{DBLP:conf/acl/LiuZDSLYJH20} is the first work to take emotional support into account in conversation. 
Meanwhile, they construct an Emotional Support Conversation dataset (ESConv) annotated with support strategies 
for the ESC task.
% with rich support strategies in a help-seeker and supporter mode. 
Based on ESConv, MISC \cite{DBLP:conf/acl/TuLC0W022} uses an off-the-shelf commonsense model COMET \cite{bosselut-etal-2019-comet} to infer an instant mental state of seeker and encodes them additionally. When predicting strategy, they take the probability of each predicted strategy as weight to get a weighted average representation of strategy, and utilize it for guiding a skillful generation.
% As each sample in the ESConv dataset also contains a brief situation summarizing the condition of the seeker, 
GLHG \cite{DBLP:conf/ijcai/00080XXSL22} considers the hierarchical relationship between the seeker's global situation (summarizing the condition of the seeker) and the local intention (inferred by COMET in each dialogue round) in conversation, and uses a graph neural network to encode their relationship for guiding generation.  
% However, most existing methods only use the limited information in current conversation and learned model parameters to provide support conversation, but ignore the powerful guidance ability of these exemplars. Since human conversation is highly subjective and open-ended, merely relying on learned model parameters is not necessarily enough, tending to generate dull and generic response.
% However, all of them are limited to the scope of current conversation and rely on learned model parameters to make a response, and they neglect a trait of support strategy, i.e. responses with the same strategy tend to be similar in expression manner. Thus, they suffer two problems: 1) barely focusing on current conversation is deficient for global perspective from historical conversation, and 2) without the guidance of expression approach, they may make an inaccurate response for target strategy. In contrast, our work sufficiently exploit exemplars to provide more problem-related and strategy-related support.
Note that both MISC and GLHG are constrained by the external knowledge in COMET, which may not be applicable to some specific domain. The external knowledge base like COMET also requires significant human effort to develop.
Meanwhile, all of them are limited to the scope of current conversation but ignore abundant prior knowledge existing in the dataset.
% This makes them limited to the scope of current conversation and just rely on learned model parameters to make a response. 
% and they neglect a trait of support strategy, i.e. responses with the same strategy tend to be similar in expression manner. Thus, they suffer two problems: 1) barely focusing on current conversation is deficient for global perspective from historical conversation, and 2) without the guidance of expression approach, they may make an inaccurate response for target strategy. 
In contrast, we focus on exploring the existing knowledge and the characteristics of the ESC task without using any external knowledge.
% In contrast, our work explore the prior knowledge in the dataset to provide problem-related and strategy-related support without using any external knowledge. 

\begin{figure*}[!t] %H为当前位置，!htb为忽略美学标准，t为浮动图形
  \centering %图片居中
  \includegraphics[width=0.85\textwidth]{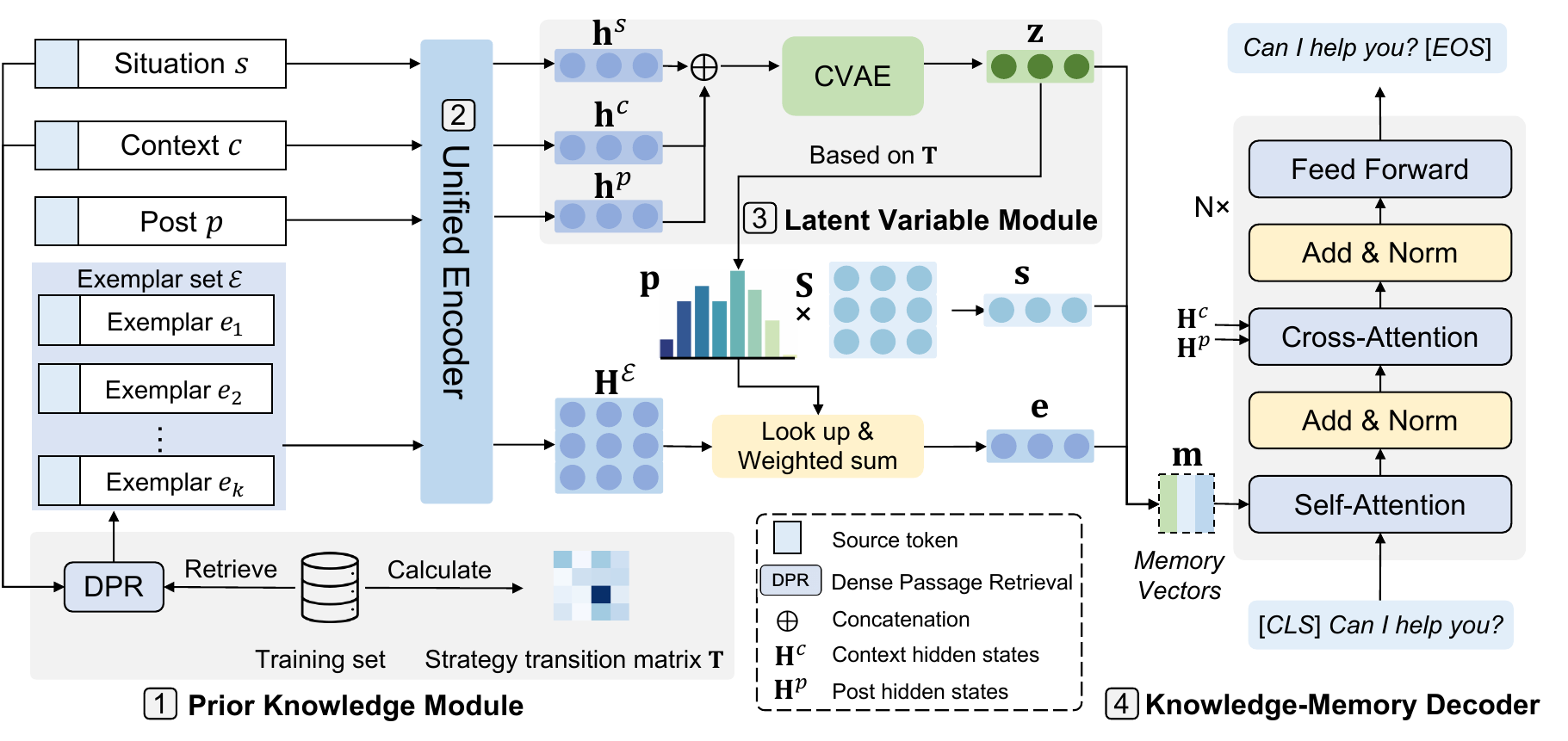} %插入图片，中设置图片大小，{}中是图片文件名
  \caption{The model architecture of PoKE, which consists of four parts: (1) Prior Knowledge Module to retrieve context-related exemplary and construct a Markov transition matrix of strategy, (2) Unified Encoder to encode multiple input sources and exemplars, (3) Latent Variable Module to sample latent variable for modeling the distribution of strategy and denoising the exemplars (using a Look up \& Weighted sum module) and (4) Knowledge-Memory Decoder to incorporate encoded prior knowledge and latent variable into the decoder.}  \label{fig:method}
  \label{fig:model}
  % \vspace{-2mm}
\end{figure*}

\subsection{Retrieve-based Generation}
There are lots of works for retrieve-based generation. We will detail some classical studies since our main aim is not to compare with them.
Some generative models, like GPT2  \cite{radford2019language}, 
perform well on many tasks such as machine translation and question answer  \cite{yang2020towards, lewis2018generative, guo2018dialog}. 
However, recent some works have pointed out that in dialogue system, the generation model just relied on the input context suffers from some issues, such as dull generation (e.g. ``I don't know") and hallucination  \cite{DBLP:conf/acl/0088GWYQWHZMCCD22, DBLP:conf/naacl/LiGBGD16, DBLP:conf/emnlp/0001PCKW21}.
To prompt model to generate more engaging response, RetNRef  \cite{DBLP:conf/emnlp/WestonDM18} proposes a simple but effective retrieve-and-refine strategy. RetNRef appends the retrieved context-relevant responses to context to guide the generation. Similar to this approach, Cai et al.  \cite{DBLP:conf/ijcai/CaiCSZY20} retrieves both literally-similar and topic-related exemplars to guide dialogue generation. 
Majumder et al.  \cite{DBLP:journals/access/MajumderGHGMP22} employs dense passage retrieval and introduce three communication mechanisms of empathy to facilitate the generation towards empathy. 
For the ESC task, the abundant prior knowledge in historical conversations has great reference value for reducing seeker's emotional stress. Besides, the responses with the same strategy are similar in sentence pattern.
Thus, we introduce exemplars into generation model and denoise exemplars according to the strategy distribution to emphasize those more relevant exemplars.

\subsection{One-to-Many Problem}
It is well known that dialogue systems have a one-to-many mapping problem that given a single context, there exist multiple valid responses  \cite{DBLP:conf/acl/0088GWYQWHZMCCD22}. To model this one-to-many feature and improve the diversity of generations, many works introduce latent variable to model a probability distribution over the potential responses  \cite{DBLP:conf/emnlp/ZengWL19, DBLP:conf/acl/ZhaoZE17, gu2018dialogwae, fang2019implicit}. 
DialogVED  \cite{DBLP:conf/acl/0088GWYQWHZMCCD22} combines continuous latent variable into the encoder-decoder pre-training framework to generate more relevant and diverse responses. 
Except for continuous representation of latent variables, some works utilize discrete categorical variables to promote the interpretability of generation  \cite{bao-etal-2020-plato, bao2021plato}.
For ESC, there also exist several reasonable support strategies and the corresponding responses at a certain stage. 
Therefore, it is required to additionally consider the one-to-many mapping relationship of strategies. 
In our work, we introduce a continuous latent variable to model the distribution over strategy. Furthermore, we employ this strategy distribution to denoise the exemplars at the sequence-level to focus on strategy-relevant exemplars.

%=======================================
\section{PoKE}
%=======================================

% \subsection{Problem Definition}
% \paragraph{\textbf{Problem Definition}}
\textbf{Problem Definition.}
% The task of emotional support conversation can.
The dialogue context in ESC is an alternating set of utterances from seeker and supporter.
Given a sequence of $N$ context utterances $ c = (u_1, u_2, \cdots, u_{N})$, where each utterance consists of some words, $u_i = (w^i_1, w^i_2, \cdots, w^i_{M})$. 
% Note that only utterance $u_i$ of supporter is annotated with a support strategy $y_{u_i} \in [0, 8)$ (support strategies are listed in Table 1).
In the setting of ESC, each utterance of supporter is labeled with a support strategy. There are total 8 support strategies, i.e. 
\textit{Question}, 
\textit{Reflection of feelings}, 
\textit{Information}, 
\textit{Restatement or Paraphrasing}, 
\textit{Others}, 
\textit{Self-disclosure}, 
\textit{Affirmation and Reassurance}, and
\textit{Providing Suggestions} (for more detail please refer to original paper  \cite{DBLP:conf/acl/LiuZDSLYJH20}).
We use $m$ to denote the total number of strategies in the following parts.
Except for the strategy, there is a brief situation $s$ ahead of conversation summarizing the condition of seeker.
% , and we denote the last one support strategy in dialogue context as $y_l$. 
In this paper, we denote the previous one support strategy taken by supporter as $y^\prime$, and the last utterance of seeker (called post) as $p$.
% For each conversation, there is also a brief situation $s$ summarizing the condition of seeker.
% , and we denote the seeker's last utterance as $x$. 
Then, our model aims at using multiple input information and prior knowledge to generate an emotional support response $r$ by reasonable support strategies.

\noindent \textbf{PoKE Overview.} Our devised model uses BlenderBot-small  \cite{roller2021recipes} as the backbone. The overview of our method is shown in Figure \ref{fig:method}, which consists of four main parts: (a) \textbf{prior knowledge module} to retrieve context-related exemplary responses and build a Markov transition matrix of strategy sequence from training set, (b) \textbf{unified encoder} to encode multiple input source and exemplars by adding source tokens, (c) \textbf{latent variable module} to model the probability distribution of strategy and denoise the exemplars and (d) \textbf{knowledge-memory decoder} to effectively incorporate encoded prior knowledge and latent variable into decoder for generation.

%=======================================
\subsection{Prior Knowledge Module}
\label{sec:prior}
%=======================================

Humans tend to use prior knowledge to bias decisions  \cite{hansen2012}, and there is abundant prior knowledge in historical conversation for ESC task. 
Due to the characteristics of ESC, we consider the prior knowledge of context-related exemplars and the general selection order of support strategies in our work. 
% The retrieved exemplary responses provide prior knowledge of expressing strategy correctly and exploring seeker's problem when coming across the similar situation again. The general selection order of support strategies supplies important cues towards making reasonable decision about strategies.

% \subsubsection{Exemplary Responses}
\noindent \textbf{Exemplary Responses.}
We use Dense Passage Retrieval (DPR)  \cite{karpukhin2020dense} as our retriever, which is a dense embedding retrieval model pre-trained on Wikipedia dump.
For a target dialogue context with the situation, DPR retrieves a set of possible supporter's responses from training set as \textit{exemplars}.
These exemplars have analogous context and situation to the current conversation.

Given the target context $c_q$ with situation $s_q$, we concatenate them as the \textit{query} input $q = [c_q, s_q]$. For each candidate response $r_p$, we get its situation $s_p$ and do the same concatenation operation to get the \textit{candidate} input $p = [r_p, s_p]$. Then, DPR calculates the similarity between the query and candidate input using the dot product of their embeddings:
\begin{equation}
  \operatorname{sim}(q, p) = E_Q (q)^T E_P (p),
\end{equation}
where $E_Q (\cdot)$ and $E_P (\cdot)$ are the encoders of query and candidate input respectively. 
In the end, we select top $k$ candidate responses with the highest similarity as exemplar set $\mathcal{E} = \{e_1, e_2, \cdots, e_k\}$, where $e_i$ denote an exemplar response. Meanwhile, we can get the corresponding strategy set $\mathcal{Y} = \{y_1, y_2, \cdots, y_k\}$, where $y_i$ denotes the strategy label of $e_i$. 
As for inference, we use the candidate encoder $E_P (\cdot)$ to pre-compute embeddings of all responses in training set, thus to save the retrieval time of inference. 
To adapt DPR to the ESC task, we fine-tune DPR on the dataset of ESC.

% \subsubsection{First-Order Markov Model of Strategy Transition} \label{sec:markov}
\noindent \textbf{First-Order Markov Model of Strategy Transition}
Before making a response, supporter need to think about reasonable strategies at different conversation stage. 
As pointed in  \cite{DBLP:conf/acl/LiuZDSLYJH20}, supporters generally follow the procedure of three stages (\textit{Exploration}, \textit{Comforting} and \textit{Action})  \cite{hill2009helping} to determine the current strategy. Thus, the general strategy order calculated in the training set can serve as prior knowledge to help decide the current strategy. 
In our work, to urge the model to focus on the previous strategy that has been chosen, we make a simple but effective assumption that the strategy sequence follows Markov chain.
% i.e. the selection of current strategy is only determined by the previous strategy. 
Then we calculate a first-order Markov transition matrix $\mathbf{T} \in \mathbb{R}^{(m+1) \times m}$ of strategy from training set, which also considers the case of no previous strategy.
Experiment in Section \ref{sec:ablation} demonstrates this is a simple but practical prior knowledge of strategy transition. The calculated strategy transition matrix is shown in Appendix \ref{app:markov}, which is used in Section \ref{sec:cvae} to help model strategy distribution.

% so they are greatly influenced by the previous strategy that has been chosen. Thus, we consider the strategy transition as a simple but effective prior knowledge to tips the selection of strategy. Con
% The strategy to be decided is determined by the conversation stage, dialogue context and the historical sequence of strategies. The

% the support strategies in ESC follow the procedure of three stages in [Hill], the general order of strategies can act as prior knowledge to help decide the current strategy, 

% \begin{table}[t]
%   \caption{Source tokens for different sources} 
%   \label{tb:source}
%   \begin{tabular}{cc}
%     \toprule
%     Source & \textit{Source token}\\
%     \midrule
%     Context & [\textit{CTX}] \\
%     Post & [\textit{POST}] \\
%     Situation & [\textit{ST}] \\
%     Exemplar & [\textit{EXEM}] \\
%   \bottomrule
% \end{tabular}
% \end{table}

%=======================================
\subsection{Unified Encoder}
%=======================================

We use a multi-layer Transformer-based encoder of BlenderBot  \cite{roller2021recipes} to encode multiple information source, including dialogue context, seeker's post, situation and the retrieved exemplars. 
Note that there are multiple source sequences to consider, so building a parameter-isolating encoder for each source will increase parameters and make training time-consuming. 
To solve this issue, we design a unified encoder, which is parameter-sharing but prepends a unique \textit{source token} to each input. \textit{Source token} can act as prompt to distinguish different input. 
There are four \textit{Source token}s including \textit{[CTX]}, \textit{[POST]}, \textit{[ST]} and \textit{[EXEM]} representing context, post, situation and exemplar, respectively.

Firstly, we reconstruct the dialogue context by concatenating them with a special token $[SEP]$ and prepending the \textit{source token} $[CTX]$, i.e. 
$c = [[CTX], w^1_1, w^1_2, \cdots, [SEP], w^2_1, w^2_2, \cdots, w^{N}_{M}]$.
 Then, we feed this sequence into the encoder to get its contextualized hidden states: 
% 统一 concatenation 操作符号。
\begin{equation} \label{eq:H}
  \mathbf{H}^c = \operatorname{Enc}(c),
\end{equation}
where $\operatorname{Enc}(\cdot)$ denotes the encoder, and $\mathbf{H}^c \in \mathbb{R}^{l \times d_h}$ is the hidden states of context sequence with $l$ tokens and hidden size of $d_h$. 
% To obtain the overall context representation of entire sequence, 
To obtain a single sentence-level representation of context, 
we take the first one hidden state of sequence, i.e. the output hidden state of \textit{source token}, as the context representation:
% The first one token in sequence, i.e. the \textit{source token}, is taken as the overall context representation:
\begin{equation}  \label{eq:h}
  \mathbf{h}^c = \mathbf{H}^c_0.
\end{equation}
% which contains rich contextual representation of the entire sequence.

Similarly, for the given situation $s$, seeker's post $p$, and each exemplar sequence $e_i$ in exemplars set $\mathcal{E} = \{e_i\}^k_{i=1}$, we prepend them with the corresponding \textit{source token} in the same way, and use the encoder to obtain their sequence representations:
\begin{align}
  \mathbf{H}^s = \operatorname{Enc}(s), \; & \mathbf{h}^s = \mathbf{H}^s_0; \nonumber\\
  \mathbf{H}^p = \operatorname{Enc}(p), \; & \mathbf{h}^p = \mathbf{H}^p_0; \nonumber\\
  \mathbf{H}^{e_i} = \operatorname{Enc}(e_i), \; & \mathbf{h}^{e_i} = \mathbf{H}^{e_i}_0,
\end{align}
% The entire representation of exemplars set $\mathcal{E}$ is expresses as:
% \begin{equation}
%   \mathbf{H}^{\mathcal{E}} = [\mathbf{h}^{e_1}, \mathbf{h}^{e_2}, ..., \mathbf{h}^{e_k}].
% \end{equation}
and we use $\mathbf{H}^{\mathcal{E}} = [\mathbf{h}^{e_1}, ..., \mathbf{h}^{e_k}]$ to express the representation of the entire exemplars set $\mathcal{E}$.
These representations of multiple source are used to model the latent variable in Section \ref{sec:cvae} and fed into the decoder for generation in Section \ref{sec:decoder}.

%=======================================
\subsection{Latent Variable Module} 
\label{sec:cvae}
%=======================================

% 总述: 
In this section, we introduce the workflow of modeling latent variable and how to build strategy distribution to obtain representations of mixed strategy and denoised exemplars.

% \subsubsection{Latent Variable} \label{sec:cvae}
\noindent \textbf{Latent Variable.} \label{sec:cvae}
% To model the one-to-many mapping relationship between context and support strategy as well as the responses, 
To address the one-to-many mapping issues of responses and support strategy at the same time, 
we utilize the Conditional Variational Autoencoder (CVAE)  \cite{sohn2015learning} to model the latent variable. 
% The graphical model of probabilistic relationships among input conditions, latent variable and response are elaborated in Figure , which shows that the dialogue context $c$, situation $s$, and seeker's post $p$  are jointly used as the input conditions for estimating the latent variable $\mathbf{z} \in \mathbb{R}^{d_z}$. 
The basic idea of CVAE is to encode the response $r$ along with input conditions to a probability distribution instead of a point. Then, CVAE employs a decoder to reconstruct the response $r$ by using latent variable $z$ sampled from the distribution. 
% During inference, CVAE samples a latent variable $z$ from the well-learned distribution, and then generates response via decoder network.
We jointly use dialogue context $c$, situation $s$, and seeker's post $p$ as the input conditions for estimating the latent variable $\mathbf{z} \in \mathbb{R}^{d_z}$. 
For brevity, we use a symbol $x = \{c, s, p\}$ to denote the input conditions. 

% In general, the latent variable is assumed to obey multivariate Gaussian distribution with a diagonal covariance matrix. Then 
CVAE is trained by maximizing a variational lower bound $\mathcal{L}_{ELBO}$, consisting of two terms: negative likelihood loss of decoder and K-L regularization: 
% \begin{align} \label{eq:cvae}
%   \mathcal{L}_{ELBO}(\theta, \phi ; r, x) &= \mathcal{L}_{nll} + \mathcal{L}_{kl} \nonumber \\
%   &=\mathbf{E}_{q_\phi(\mathbf{z} | x, r)}\left[\log p_\theta(r | \mathbf{z}, x)\right]  \nonumber \\
%   &-K L\left(q_\phi(\mathbf{z} | r, x) \| p_\theta(\mathbf{z} | x)\right) \nonumber \\
%   & \leq \log p(r | x),
% \end{align}
\begin{align} \label{eq:cvae}
  \mathcal{L}_{ELBO} &= \mathcal{L}_{nll} + \mathcal{L}_{kl}  \\
  % &=\mathbf{E}_{q_\phi(\mathbf{z} | x, r)}\left[\log p_\theta(r | \mathbf{z}, x)\right] -K L\left(q_\phi(\mathbf{z} | r, x) \| p_\theta(\mathbf{z} | x)\right) \nonumber 
  &=\mathbf{E}_{q_\phi(\mathbf{z} | x, r)}\left[\log p_\theta(r | \mathbf{z}, x)\right]  \nonumber \\
  &-K L(q_\phi(\mathbf{z} | r, x) \| p_\theta(\mathbf{z} | x)) \nonumber 
  % & \leq \log p(r | x),
\end{align}
where $q_\phi(\mathbf{z} | r, x)$ and $p_\theta(\mathbf{z} | x)$ are called recognition network and prior network respectively (with parameters $\phi$ and $\theta$), and $p_\theta(r | \mathbf{z}, x)$ is the decoder for generation, which will be illustrated in Section \ref{sec:decoder}. Then we can sample latent variable $\mathbf{z}$ from the well-learned Gaussian distribution (for more detail please see Appendix \ref{app:cvae}).

% The basic idea of VAE is to encode the input x into a probability distribution z instead of a point encoding in the autoencoder. Then VAE applies a decoder network to reconstruct the original input using samples from z. To generate images, VAE ﬁrst obtains a sample of z from the prior distribution, e.g. N (0, I), and then produces an image via the decoder network.
% A more advanced model, the conditional VAE (CVAE), is a recent modiﬁcation of VAE to generate diverse images conditioned on certain attributes, e.g. generating different human faces given skin color

In order to regularize the latent space and model the one-to-many mapping relationship of strategy, we design an extra optimizing objective of strategy, $\mathcal{L}_y$. A strategy prediction network $p_\theta(y | \mathbf{z})$ is used to recover the strategy label $y$ by latent variable $\mathbf{z}$:
\begin{align} \label{eq:strategy}
  \mathcal{L}_y = \mathbf{E}_{q_\phi(\mathbf{z} | x, r)}& [p_\theta(y | \mathbf{z})], \\
  % p_\theta(y | \mathbf{z}) &= \mathbf{W}_y \mathbf{z} + \mathbf{b}_y + \mathbf{T}(y)
  p_\theta(y | \mathbf{z}) &= \mathbf{p}_y,  
  % \mathbf{p} &= \mathbf{W}_y \mathbf{z} + \mathbf{b}_y + \mathbf{T}(y)
\end{align}
where $\mathbf{p}$ is denoted as the distribution of strategy. We calculate $\mathbf{p}$ by a fully connected layer and based on the transition matrix $\mathbf{T}$ obtained in Section \ref{sec:prior}:
\begin{equation} \label{eq:distribution}
  \mathbf{p} =  \text{softmax}(\mathbf{W}_y \mathbf{z} + \mathbf{b}_y + \mathbf{T}_{y^\prime}),
\end{equation}
where $\mathbf{W}_y \in \mathbb{R}^{m \times d_z}$ and $\mathbf{b}_y \in \mathbb{R}^{m}$ are the learnable parameters, $y^\prime$ is the previous strategy taken by the supporter, which is provided in dataset, and $\mathbf{T}_{y^\prime} \in \mathbb{R}^{m}$ is the transition probability of $y^\prime$. 

% \subsubsection{Representation of Mixed Strategy}
\noindent \textbf{Representation of Mixed Strategy.}
To model the complexity of strategy expressed in one utterance, and consider multiple valid support strategies, we adopt a method of mixed strategy representation inspired by  \cite{DBLP:conf/acl/TuLC0W022, DBLP:conf/emnlp/LinMSXF19}.
% we utilize the strategy distribution $\mathbf{p}$ and 
First, we create a strategy codebook $\mathbf{S} \in \mathbb{R}^{m \times d_h}$ storing the representation of each strategy. Then, we utilize the strategy distribution $\mathbf{p}$ to get a weighted combination of $\mathbf{S}$, which blends multiple strategy in one representation $\mathbf{s} \in \mathbb{R}^{d_h}$:
\begin{equation}
  \mathbf{s} = \mathbf{p} \cdot \mathbf{S}.
\end{equation}

% \subsubsection{Representation of Denoised Exemplars}
\noindent \textbf{Representation of Denoised Exemplars.}
% As illustrated in introduction
% In general, the retrieved exemplars contain different support strategies,
In general, the retrieved exemplars contain irrelevant support strategies.
% \textit{question}, \textit{self-disclosure} etc., 
% and some strategies irrelevant with the target strategy are noisy. Using them without denoising would mislead the expression approach of target strategy. 
To denoise the exemplars in terms of strategy, we first look up the strategy probability from strategy distribution as the weight for each exemplar. Then, we combine all exemplar representations $\mathbf{H}^{\mathcal{E}}$ at sequence level to obtain a single representation $\mathbf{e} \in \mathbb{R}^{d_h}$ of denoised exemplars:
\begin{align}\label{eq:denoise}
  % \mathbf{e} = \sum^k_{i=1} \frac{\mathbf{p}[y_i]}{\sum^k_{i=1} \mathbf{p}[y_i] }\cdot \mathbf{H}^{\mathcal{E}}_i ,
  \mathbf{e} &= \sum^k_{i=1} \frac{\mathbf{p}_{y_i}}{\sum^k_{j=1} \mathbf{p}_{y_j} }\cdot \mathbf{H}^{\mathcal{E}}_{i} \\ \nonumber
  &= \sum^k_{i=1} \frac{\mathbf{p}_{y_i}}{\sum^k_{j=1} \mathbf{p}_{y_j} }\cdot \mathbf{h}^{e_i} ,
\end{align} 
where $y_i \in [0, m)$ is the strategy label of exemplar $e_i$, $\mathbf{p}_{y_i}$ denotes the probability of $y_i$, and the denominator is normalization.
% , and $\mathbf{H}^{\mathcal{E}}_i$ denote the labeled strategy and sequence representation of the $i$th exemplar $e_i$. 

The representations of the latent variable, mixed strategy, and denoised exemplars will be incorporated into the decoder to guide generation, which is illustrated in the following section.

\begin{figure}[h] 
  \centering
  \includegraphics[width=0.35\linewidth]{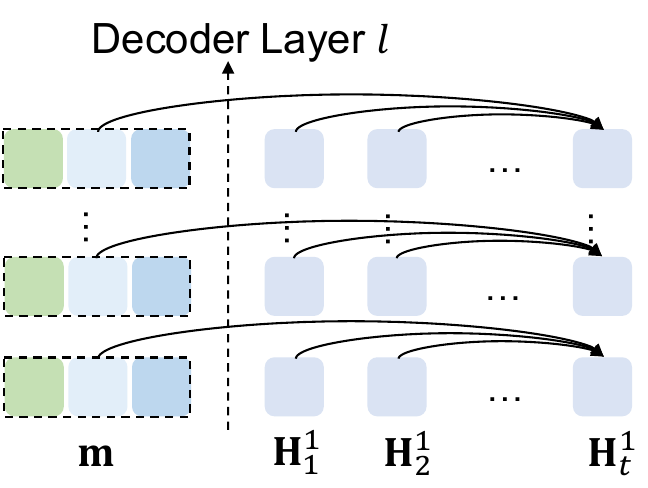}
  \caption{Illustration of memory schema applied in self-attention module in decoder: $\mathbf{H}^{l}_{t}$ attends both $\mathbf{H}^{l}_{<t}$ and \textit{memory} vectors $\mathbf{m}$ at each layer.} \label{fig:memory}
  \label{fig:example}
  % \vspace{-3mm}
\end{figure}

\subsection{Knowledge-Memory Decoder} \label{sec:decoder}

% To effectively inject the latent variable, and the representations of mixed strategy as well as denoised exemplars into the decoder, we 
After getting the above-mentioned representations of the latent variable $\mathbf{z}$, mixed strategy $\mathbf{s}$, and denoised exemplars $\mathbf{e}$, the consequent problem is how to effectively incorporate them into decoder\footnote{We apply the decoder in BlenderBot \cite{roller2021recipes} to model the distribution $p_\theta(r | \mathbf{z}, x)$ and optimize the negative likelihood loss $\mathcal{L}_{nll}$ in Eq. (\ref{eq:cvae}). } for generation.
% To effectively inject these knowledge of 
% one-to-many mapping relationship, 
% latent variable, mixed strategies and denoised exemplars into decoder, we refine the decoder with a memory schema  \cite{DBLP:conf/acl/0088GWYQWHZMCCD22, DBLP:conf/emnlp/LiGLPLZG20}.
Inspired by \cite{DBLP:conf/acl/0088GWYQWHZMCCD22, DBLP:conf/emnlp/LiGLPLZG20}, we apply a memory schema to inject these encoded knowledge.
The memory schema regards the representations of the encoded knowledge as additional \textit{memory vectors}  $\mathbf{m}$ for each self-attention layer to attend, as illustrated in Figure \ref{fig:memory}. 
% Specifically, we first calculate \textit{memory} vector of latent variable by projecting it into the $d_h$-dimensional space:
We first project the vector of latent variable $\mathbf{z}$ into the $d_h$-dimensional space:
\begin{equation}
  \mathbf{z}_h = \mathbf{W}_z \mathbf{z},
\end{equation}
where $\mathbf{W}_z \in \mathbb{R}^{d_h \times d_z}$ is the projection matrix. 
Thus, we can obtain the memory vectors $\mathbf{m} = [\mathbf{z}_h, \mathbf{s}, \mathbf{e}] \in \mathbb{R}^{3 \times d_h}$ by stacking $\mathbf{z}_h, \mathbf{s}, \mathbf{e}$.
% Then, we modify the key $K$ and value $V$ in each self-attention layer by prepending \textit{memory} vectors $\mathbf{m} = [\mathbf{z}_h, \mathbf{s}, \mathbf{e}] \in \mathbb{R}^{3 \times d_h}$:
Then, we modify the computation of key vector $K$ and value vector $V$ in each self-attention layer by incorporating the memory vectors. Concretely, memory vectors $\mathbf{m}$ are prepended to the hidden states $\mathbf{H}^{l}$, denoted as $[\mathbf{m}, \mathbf{H}^{l}]$, to calculate the key vector $K$ and value vector $V$ in each self-attention layer:
\begin{align}
  % \operatorname{SelfAttn}(Q, K, V) = \operatorname{SelfAttn}(\mathbf{W}^Q\mathbf{H}^{l}, \mathbf{W}^K[\mathbf{m}, \mathbf{H}^{l}], \mathbf{W}^V[\mathbf{m}, \mathbf{H}^{l}]),
  % Q = \mathbf{W}^Q\mathbf{H}^{l}  \nonumber \\
  % K = \mathbf{W}^K\mathbf{H}^{l}  \\
  % V = \mathbf{W}^V\mathbf{H}^{l} \nonumber 
  % Q = \mathbf{W}^Q\mathbf{H}^{l}  ;
  K = [\mathbf{m}, \mathbf{H}^{l}]\mathbf{W}^K \nonumber  \\
  V = [\mathbf{m}, \mathbf{H}^{l}]\mathbf{W}^V 
\end{align}
where $\mathbf{W}^K, \mathbf{W}^V \in \mathbb{R}^{d_h \times d_h}$ are parameter matrices of key and value, respectively.
% , and $\mathbf{H}^{l}$ denote the output hidden states of $l$-th layer in decoder.
The memory schema is equivalent to adding some virtual tokens to the response sequence at each layer and enables the decoder to attend all knowledge directly. 
Besides, we perform multi-head attention over the encoded context $\mathbf{H}^{c}$ and post $\mathbf{H}^{p}$ for each layer's cross attention inspired by  \cite{DBLP:conf/acl/TuLC0W022}.
% For each layer's cross attention module, we follow the Transformer to performs multi-head attention over the context output $\mathbf{H}^{c}$ of the encoder stack.
In this way, the knowledge is injected into the decoder to guide the generation at each step.
% The insight of memory schema is that \textit{memory} vectors are equivalent to adding some virtual tokens to response sequence at each layer, and enables decoder to attend all knowledge directly. 

%=======================================
\subsection{Training Objective}
%=======================================

The final learning objective is defined as the combination of CVAE loss in Eq. (\ref{eq:cvae}) and strategy prediction loss in Eq. (\ref{eq:strategy})
\begin{equation} \label{eq:objective}
  \mathcal{L}(\varphi ) = \mathcal{L}_{ELBO} + \lambda\mathcal{L}_y,
\end{equation}
where $\varphi$ denotes the parameters of PoKE, and $\lambda$ controls the degree of regularizing latent space by strategy. However, directly training this objective may suffer two optimizing challenges, i.e. \textit{KL-vanishing} and \textit{strategy-unstablity}. To alleviate them, we adopt two annealing methods including \textit{KL-annealing} and \textit{Strategy-annealing}. 

\noindent \textbf{KL-vanishing.} This problem lies in that the decoder overly attends the encoded information of context, and thus ignore the latent variable $\mathbf{z}$, leading to the failure of encoding informative $\mathbf{z}$  \cite{DBLP:conf/conll/BowmanVVDJB16}. 
We adopt a \textit{KL annealing}  \cite{DBLP:conf/acl/ZhaoZE17} method to solve this issue, i.e. gradually increasing the weight of KL loss in Eq. (\ref{eq:cvae}) from 0 to 1 during train.

\begin{table*}[t]
  \caption{Result of automatic evaluation on baseline models and PoKE. $^{*}$ denotes the model requiring external knowledge. The best performance under the setting of no external knowledge is highlighted in \textbf{bold}. Considering the model using external knowledge, the best score is \underline{underlined}. $\downarrow$ indicates that the lower the value, the better the performance.  } \label{tb:main}
  \begin{tabular}{lcccccccc}
    \toprule
    \textbf{Model}            & \textbf{PPL} $\downarrow$   & \textbf{B-1} $\uparrow$  & \textbf{B-2} $\uparrow$  & \textbf{B-3} $\uparrow$  & \textbf{B-4} $\uparrow$ & \textbf{R-L} $\uparrow$ & \textbf{D-1} $\uparrow$   & \textbf{D-2} $\uparrow$  \\
    \midrule
    \rowcolor{lightgray!40}
    \multicolumn{9}{l}{\textbf{w/o external knowledge}}\\
    Transformer      &  53.85  & 15.07 & 4.67 & 1.78 & 0.84 & 13.26  & 1.49 &  12.97 \\
    MultiTRS         &  53.08 & 15.06  & 4.67 &  1.74 & 0.77&  13.45 & 1.56 &  13.65 \\
    MoEL             &  53.61 & 17.98  &  5.96& 2.27 & 1.02 &  14.08 & 1.12 & 11.25 \\
    % BlenderBot        & 15.92 & 17.75  & 6.49 & 3.09     & 1.72 & 15.56 & 3.46 & 20.97 \\
    BlenderBot-Joint  & \underline{\textbf{15.71}} &  16.99  & 6.18 & 2.95 & 1.66 & 15.13 & 3.27 & 20.87 \\
    \midrule
    \rowcolor{lightgray!40}
    \multicolumn{9}{l}{\textbf{with external knowledge}}\\
    MISC$^{*}$              & 16.62 & 17.71  & 6.44 & 3.00  & 1.62 & 15.57 & 3.65 & \underline{22.25} \\
    \midrule
    PoKE             & 15.84 & \underline{\textbf{18.41}} & \underline{\textbf{6.79}} & \underline{\textbf{3.24}} & \underline{\textbf{1.78}} & \underline{\textbf{15.84}} & \underline{\textbf{3.73}} & \textbf{22.03} \\
    \bottomrule
  \end{tabular}
\end{table*}

\noindent \textbf{Strategy-unstablity.} At the early stage of training, using latent variable tends to predict unstable and incorrect strategy distribution. Then, this error is propagated to the representation of denoised exemplars and the decoder  \cite{DBLP:conf/emnlp/LinMSXF19}. To stabilize the training stage, we take a measure of \textit{strategy-annealing}. That is, we use the true distribution of target strategy instead of the predicted distribution by a certain probability $\alpha_t$ and anneal it over time:
\begin{equation}
  \alpha_t = \beta + (1 - \beta)e^{- \frac{t}{T}}
\end{equation}
where $\beta$ is annealing rate, $t$ is the current iteration step, and $T$ is the annealing steps.

\begin{table}[ht]
  \caption{Statistics of processed split ESConv.} 
  \label{tb:split}
  \begin{tabular}{lccc}
    \toprule
    \textbf{Category} & \textbf{Train} & \textbf{Valid} & \textbf{Test} \\
    \midrule
    \# Dialogues  & 12,235  & 2,616 & 2,794 \\
    Avg. length of turns  & 8.57 & 8.58  & 8.65 \\
    Avg. length of utterances & 18.34  & 18.31 & 17.04 \\
    Avg. length of contexts & 157.35 & 157.18 & 147.52 \\
    % Avg. length of situations & 23.24  & 22.86 & 22.37 \\
    % Avg. length of posts & 15.21  & 15.67 & 14.66 \\
    \bottomrule
  \end{tabular}
  % \vspace{-4mm}
\end{table}

\section{Experiments}
% \subsection{Experimental Configurations}
% \subsubsection{Dataset}
\subsection{Dataset}
We use the emotional support conversation dataset \textbf{ESConv}  \cite{DBLP:conf/acl/LiuZDSLYJH20} to evaluate our method. 
ESConv contains a total of 1,053 dialogues and 31,410 utterances.
Each conversation contains a seeker's situation and a dialog context, and each utterance of supporter is annotated by a support strategy that is taken by the supporter. There are 8 different support strategies roughly uniformly distributed across the dataset. Due to the long turns in ESC, we cut each conversation into several pieces with 10 utterances and the last utterance is supporter's response. For training and validation, we split the ESConv into the sets of training/validation/test with the proportions of 7:1.5:1.5. 
The statistics of original ESConv is shown in Table \ref{tb:esconv} and the split ESConv is shown in Table \ref{tb:split}. 

\subsection{Evaluation Protocol}
Following existing methods, we adopt automatic and human evaluation to evaluate our model and compare with strong baselines.

% \paragraph{Automatic Evaluation} 
\noindent \textbf{Automatic Evaluation.}
We employ perplexity (PPL), BLEU-1 (B-1), BLEU-2 (B-2), BLEU-3 (B-3), BLEU-4 (B-4)  \cite{DBLP:conf/acl/PapineniRWZ02}, ROUGE-L (R-L)  \cite{lin2004rouge}, Distinct-1 (D-1), Distinct-2 (D-2)  \cite{li-etal-2016-diversity} automatic metrics to evaluate model performance. 
PPL is defined as $e$ raised to the power of cross-entropy and is kept as a reference. B-1/2/3/4 and ROUGE-L measure the number of matching n-grams between the model-generated response and the human-produced reference, which reflect the quality generation. D-1/2 is calculated by the number of distinct 1/2-grams divided by the total number of generated words, which indicates the generation diversity.
% PPL, B-1/2/3/4, and R-L reflect the quality of generation and D-1/2 indicates the generation diversity. 

% \paragraph{Human Evaluation} 
\noindent \textbf{Human Evaluation.}
We randomly sample 64 dialogues from the test set and generate responses using our model and one baseline. Then, 3 annotators with relevant backgrounds are prompted to choose the better response based on indicators in  \cite{DBLP:conf/acl/LiuZDSLYJH20}: (1) Fluency: which one are more fluent? (2) Identification: which one is more helpful in identifying the seeker's problems? (3) Comforting: which one is more skillful in comforting the seeker? (4) Suggestion: which one provides more helpful suggestions? (5) Overall: generally, which emotional support do you prefer? 
% For a fair comparison, annotators do not know which model the response is generated from.

\subsection{Compared Methods}
% Note that our main aim is to improve the model under the setting of no external knowledge. For comparisons with the baselines, we place emphasis on those with no need for external knowledge.
% We place emphasis on those baselines that do not require external knowledge, since our main purpose is to improve the model under the setting of no external knowledge.
Since our main purpose is to explore the ESC task under the setting of no external knowledge, we place emphasis on those baselines that do not require any external knowledge.
% We compare our model with the following baselines, and one model that uses external knowledge:
We compare our model with the following baselines, also including a model using external knowledge:
% \begin{enumerate}
%   \item \textbf{Transformer}  \cite{DBLP:conf/nips/VaswaniSPUJGKP17}. We use a standard Transformer model, which is trained from scratch by negative likelihood objective.
%   \item \textbf{Multi-TRS}  \cite{DBLP:journals/corr/abs-1811-00207}. Multi-TRS is a multitask Transformer trained based on an additional supervised learning objective predicting the target emotion.
%   \item \textbf{MoEL}  \cite{DBLP:conf/emnlp/LinMSXF19}. MoEL is a model mixed of empathetic listeners, which models the distribution of emotion and assign it to multiple decoders to softly combine their output.
%   % \item \textbf{MIME}. MIME  generates empathetic response that mimic the emotion of the speaker while accounting for their affective charge (positive or negative).
%   \item \textbf{BlenderBot-Joint}  \cite{DBLP:conf/acl/LiuZDSLYJH20}. BlenderBot-Joint is proposed along with ESConv dataset, which is built on a large scale model BlenderBot. It generates a strategy token firstly and attach it to the head of response to make a desired response.
%   \item \textbf{MISC}  \cite{DBLP:conf/acl/TuLC0W022}. MISC is also built on BlenderBot. It firstly infers the user's fine-grained emotional status using COMET  \cite{bosselut-etal-2019-comet}, and then responds skillfully using a mixture of strategy.
% \end{enumerate}
\begin{enumerate}
  \item \textbf{Transformer}  \cite{DBLP:conf/nips/VaswaniSPUJGKP17}. We use a standard Transformer model, which is trained from scratch by a negative likelihood objective.
  \item \textbf{Multi-TRS}  \cite{DBLP:journals/corr/abs-1811-00207}. Multi-TRS is a multitask Transformer trained with an additional learning objective of predicting the target emotion.
  \item \textbf{MoEL}  \cite{DBLP:conf/emnlp/LinMSXF19}. MoEL models the distribution of emotion and assigns it to multiple Transformer decoders to softly combine their output.
  \item \textbf{BlenderBot-Joint}  \cite{DBLP:conf/acl/LiuZDSLYJH20}. BlenderBot-Joint is built on a pre-trained dialogue model, BlenderBot \cite{roller2021recipes}. It generates a strategy token and attaches it to the head of response to guide the desired response.
  \item \textbf{MISC}   \cite{DBLP:conf/acl/TuLC0W022}. MISC is also built on BlenderBot but requires external knowledge. It injects external knowledge by inferring the user's fine-grained emotional status using COMET  \cite{bosselut-etal-2019-comet}. When generating, they first predict a probability distribution of strategy and use it to obtain a weighted average representation of strategy for guiding generation.
\end{enumerate}
% \textbf{Transformer}  \cite{DBLP:conf/nips/VaswaniSPUJGKP17}. We use a standard Transformer model, which is trained from scratch by a negative likelihood objective. \\
% \textbf{Multi-TRS}  \cite{DBLP:journals/corr/abs-1811-00207}. Multi-TRS is a multitask Transformer trained on an additional learning objective predicting the target emotion.\\
% \textbf{MoEL}  \cite{DBLP:conf/emnlp/LinMSXF19}. MoEL models the distribution of emotion and assigns it to multiple Transformer decoders to softly combine their output.\\
%   % \item \textbf{MIME}. MIME  generates empathetic response that mimic the emotion of the speaker while accounting for their affective charge (positive or negative).
% \textbf{BlenderBot-Joint}  \cite{DBLP:conf/acl/LiuZDSLYJH20}. BlenderBot-Joint is built on a pre-trained dialogue model, BlenderBot. It generates a strategy token and attaches it to the head of response to make the desired response.\\
% \textbf{MISC}  \cite{DBLP:conf/acl/TuLC0W022}. MISC is also built on BlenderBot. It injects external knowledge by inferring the user's fine-grained emotional status using COMET  \cite{bosselut-etal-2019-comet}. When generating, they first predict a probability distribution of strategy and use it to obtain a weighted average representation of strategy for guiding generation.
% %  and responds skillfully by a mixture of strategy.

Note that Multi-TRS and MoEL require the emotion label of seeker for training, so we use the conversation-level emotion label provided in ESConv dataset to train them. 
For a fair comparison, we apply the same hyperparameters for all baselines. The detail of implementation is illustrated in Appendix \ref{app:implementation} 

% \subsubsection{Implementation Details}
% Appendix?

\subsection{Experiment Results}
\noindent \textbf{Automatic Evaluation.}
The automatic evaluation results compared with baseline models are shown in Table \ref{tb:main}. The results show that PoKE significantly outperforms baselines on the majority of metrics. This indicates PoKE can generate high-quality and more diverse responses, which proves the superiority of PoKE.
%  and demonstrates the effectiveness of PoKE. 
% Although MoEL employs more complex model architecture to inject appropriate emotion into the generation, it still does not perform as well. The reason is that it only emphasizes on empathetic response but ignores the task features of emotional support conversation, i.e. reducing user's emotional stress using various support strategies with conversation goes. In a nutshell, MoEL fails to model the correlation between support strategy and generation. 

Specifically, the Transformer-based models, i.e. Transformer, Multi-TRS, and MoEL, do not perform well on ESConv. This is because these models are initialized with random parameters and trained on ESConv from scratch. Besides, their training objectives are irrelevant to the support strategy and the characteristics of emotional support, so they are hard to handle the challenging ESC task.
As for the BlenderBot-based model, i.e. BlenderBot-Joint and MISC, they gain an improvement by a large margin compared to the previous baselines. It is due to the pre-trained dialogue model BlenderBot, which is trained on a large conversation dataset containing multiple conversation skills  \cite{DBLP:conf/acl/SmithWSWB20}.
For MISC, its D-1 and D-2 are comparatively higher, indicating that it tends to generate more diverse responses. This is because MISC incorporates varied information about seeker's mental state from external knowledge in COMET and merges mixed strategies into one response. However, due to the issue of the local scope of conversation and the one-to-many relationship of strategy, there is still room for improvement.

% However, they are limited to the local limited to the scope of 
% current conversation and rely on learned model parameters to make a response.  
% ignore two significant characteristics of ESC: (a) abundant prior knowledge in historical conversations, such as the responses to similar cases and the order of support strategies, which has a great reference value for guiding supporter to explore user's problem and decide the target support strategy, and (b) the one-to-many mapping relationship that multiple strategies are valid for a single context. 

Compared to those baselines without external knowledge, our proposed model PoKE improves significantly on the majority of metrics. 
This demonstrates that by effectively exploiting global prior knowledge from historical conversations,
% of exemplary responses and strategy decision, 
PoKE can get more clues to focus on seeker's problem and generate more relevant responses.
For the MISC that uses additional external knowledge, PoKE still can obtain a slight improvement in some metrics except diversity.
% On the metrics of diversity, i.e. D-1 and D-2, PoKE 
% match or even surpass 
However, PoKE
almost achieves the same diversity performance with 
MISC.
%  even without introducing external knowledge. 
This benefits from using latent variable to model the one-to-many mapping relationship between context and support strategy, and latent variable makes it easier to sample infrequent strategies. Moreover, the technique of mixed strategy facilitates expressing diverse strategies in one response.
As for PPL, both MISC and PoKE perform worse than BlenderBot-Joint. 
% This is reasonable since better diversity generally results in higher PPL  \cite{9406340}. 
A recent work proves that PPL is not so reliable for evaluating text quality  \cite{https://doi.org/10.48550/arxiv.2210.05892}, 
and due to the insignificant difference of PPL (PoKE only drops by 0.13), we do not further refine the model.
% to save computational resources.

\begin{table}[t]
  \caption{Human evaluation results.} \label{tb:human}
  \begin{tabular}{lcccc}
    \toprule
    \textbf{Comparisons} & \textbf{Indicators} & \textbf{Win} & \textbf{Lose} & \textbf{Tie} \\
    \hline & Flu. & $\mathbf{61.0}$ & $8.2$ & $29.2$ \\
    & Ide. & $\mathbf{64.6}$ & $13.3$ & $20.5$ \\
    PoKE vs. MoEL & Com. & $\mathbf{68.7}$ & $15.3$ & $14.3$ \\
    & Sug. & $\mathbf{65.6}$ & $14.8$ & $17.9$ \\
    & Ove. & $\mathbf{70.2}$ & $14.8$ & $13.3$ \\
    \hline & Flu. & $\mathbf{30.2}$ & $23.4$ & $46.3$ \\
    & Ide. & $\mathbf{37.5}$ & $29.6$ & $32.8$ \\
    PoKE vs. MISC$^{*}$ & Com. & $\mathbf{43.2}$ & $33.3$ & $22.9$ \\
    & Sug. & $\mathbf{36.4}$ & $30.2$ & $33.3$ \\
    & Ove. & $\mathbf{45.8}$ & $34.8$ & $19.2$ \\
    \bottomrule
  \end{tabular}  
\end{table}

\noindent \textbf{Human Evaluation.} 
The best Transformer-based model MoEL and BlenderBot-based model MISC are used to do a further human evaluation, which is shown in Table \ref{tb:human}. 
% The result is shown in Table \ref{tb:human}.
The result displays that our proposed PoKE is superior to MoEL and MISC on all indicators, which is nearly consistent with the automatic evaluation results. Significantly, our PoKE outperforms MoEL by a large margin. This is partly due to the pre-trained backbone model Blenderbot, which contains abundant knowledge about communication skills. 
Compared with MISC, our PoKE that does not rely on external knowledge also achieves a decent performance, especially on aspects of Comforting and Identification.  
This indicates that 
% We speculate that 
the retrieved context-related exemplars contains a lot of information relevant to seeker's problem, which  gives model more clues to identify the current problem and comfort seeker.  

Overall speaking, under the setting of no external knowledge, our proposed PoKE is superior to baselines on both automatic evaluation and human evaluation, which proves the superiority and effectiveness of PoKE. Besides, abundant prior knowledge and latent variable help provide better and diverse emotional support in dialogue system.

% and thus gives model more clues to focus on seeker's problem and express strategy more accurately.

% For human evaluation, we consider the best Transformer-based model MoEL and BlenderBot-based model MISC.
% We compare PoKE with a Transformer-based 

\begin{table}[t]
  \caption{Analysis of denoised exemplars} \label{tb:denoise}
  \resizebox{\columnwidth}{!}{\begin{tabular}{lcccccc}  
    \toprule
    \textbf{Model}                    & \textbf{PPL} $\downarrow$   & \textbf{B-2} $\uparrow$  & \textbf{B-4} $\uparrow$  & \textbf{R-L} $\uparrow$  & \textbf{D-1} $\uparrow$  & \textbf{D-2} $\uparrow$  \\
    \midrule
    PoKE       & \textbf{15.84} & \textbf{6.79} & \textbf{1.78} & \textbf{15.84} & \textbf{3.73} & \textbf{22.03} \\
    w/o denoising          & 15.81 & 6.76 & 1.69 & 15.60 & 3.58 & 21.23 \\
    
    % Structure (c) & 15.85 & 6.53 & 1.69 & 15.25 & \textbf{3.80} & \textbf{22.67} \\
    % Model                    & PPL $\downarrow$   & B-2 $\uparrow$  & B-4 $\uparrow$  & R-L $\uparrow$  & D-1 $\uparrow$    \\
    % \midrule
    % Structure (b)          & 16.07 & 6.76 & 1.78 & 15.61 & 3.28  \\
    % Structure (c) & 15.85 & 6.53 & 1.69 & 15.25 & \textbf{3.80}  \\
    % PoKE       & \textbf{15.84} & \textbf{6.79} & \textbf{1.78} & \textbf{15.84} & 3.73  \\
    \bottomrule
    \end{tabular}}
\end{table}

% \begin{table}[t]
%   \caption{Analysis of the number of exemplars $k$} \label{tb:num exem}
%   \begin{tabular}{lcccccc}  
%     \toprule
%     % Model                    & PPL $\downarrow$   & B-2 $\uparrow$  & B-4 $\uparrow$  & R-L $\uparrow$  & D-1 $\uparrow$  & D-2 $\uparrow$  \\
%     \textbf{Model}                    & \textbf{PPL} $\downarrow$   & \textbf{B-2} $\uparrow$  & \textbf{B-4} $\uparrow$  & \textbf{R-L} $\uparrow$  & \textbf{D-1} $\uparrow$  & \textbf{D-2} $\uparrow$  \\
%     \midrule
%      $k$ = 1  &15.86	&6.71&	1.63& 15.52&	3.32	&20.24	\\
%      $k$ = 5  & 15.86	&6.76&	1.71& 15.53&	3.30	&20.41	\\
%      $k$ = 10 & \textbf{15.84}	&6.79&	\textbf{1.78}& 15.84&	\textbf{3.73}	&\textbf{22.03}	\\
%      $k$ = 15 &16.40	&\textbf{7.13}&	1.73& \textbf{15.88}&	3.26	&19.79	\\

%     % $k$ = 1       & 15.86 & 6.76 & 1.69 & 15.60 & 3.58 & 21.23 \\
%     % $k$ = 5          & 15.81 & 6.76 & 1.69 & 15.60 & 3.58 & 21.23 \\
%     % $k$ = 10          & 15.81 & 6.76 & 1.69 & 15.60 & 3.58 & 21.23 \\
%     % $k$ = 15          & 15.81 & 6.76 & 1.69 & 15.60 & 3.58 & 21.23 \\
    
%     % Structure (c) & 15.85 & 6.53 & 1.69 & 15.25 & \textbf{3.80} & \textbf{22.67} \\
%     % Model                    & PPL $\downarrow$   & B-2 $\uparrow$  & B-4 $\uparrow$  & R-L $\uparrow$  & D-1 $\uparrow$    \\
%     % \midrule
%     % Structure (b)          & 16.07 & 6.76 & 1.78 & 15.61 & 3.28  \\
%     % Structure (c) & 15.85 & 6.53 & 1.69 & 15.25 & \textbf{3.80}  \\
%     % PoKE       & \textbf{15.84} & \textbf{6.79} & \textbf{1.78} & \textbf{15.84} & 3.73  \\
%     \bottomrule
%     \end{tabular}
% \end{table}

\subsection{Effect of Exemplars} \label{sec:exem}
In this section, we explore the effect of exemplars in terms of denoising and quantity.
To verify the denoised exemplars in Eq. (\ref{eq:denoise}), we implement another variant of PoKE without denoising exemplars, i.e. the representation of exemplars is calculated by averaging, i.e. $\mathbf{e} = \frac{1}{k}\sum^k_{i=1} \mathbf{h}^{e_i}$. The result is displayed in Table \ref{tb:denoise}. All metrics drop when not denoising the exemplars. This demonstrates that the strategies of some retrieved exemplars are irrelevant to the current context, and need to be used selectively.

\begin{figure}[t] 
  \centering
  \includegraphics[width=\linewidth]{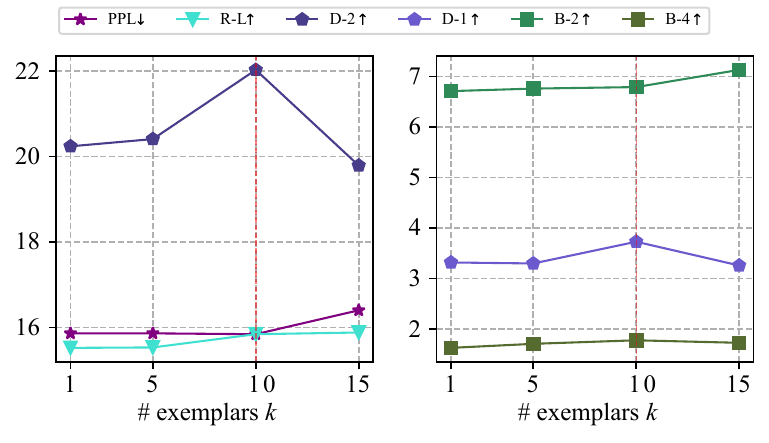}
  \caption{Analysis of the number of exemplars $k$} \label{fig:num exem}
\end{figure}

Figure \ref{fig:num exem} shows that as the number of exemplars increases, the overall performance tends to improve first and then decrease. 
This is because when exemplars are insufficient, PoKE lacks adequate reference information. When exemplars are too many, there is a lot of redundant and noisy information to distract the generation. 
Although PoKE ($k$ = 15) can utilize plentiful information to improve quality (higher B-2 and R-L), it pays the price of decreased fluency and diversity (very low PPL and D-1/2).
In the end, we decide to retrieve 10 exemplars for each sample considering both the overall effect and training efficiency.

%  our proposed SNRI is capable of modeling complete neighboring relations to handle sparse subgraphs.

\begin{table}[t]
  \caption{The results of PoKE with different CVAE structure.} \label{tb:cvae}
  \resizebox{\columnwidth}{!}{\begin{tabular}{lcccccc}  
    \toprule
    % Model                    & PPL $\downarrow$   & B-2 $\uparrow$  & B-4 $\uparrow$  & R-L $\uparrow$  & D-1 $\uparrow$  & D-2 $\uparrow$  \\
    \textbf{Model}                    & \textbf{PPL} $\downarrow$   & \textbf{B-2} $\uparrow$  & \textbf{B-4} $\uparrow$  & \textbf{R-L} $\uparrow$  & \textbf{D-1} $\uparrow$  & \textbf{D-2} $\uparrow$  \\
    \midrule
    Normal CVAE       & \textbf{15.84} & \textbf{6.79} & \textbf{1.78} & \textbf{15.84} & \textbf{3.73} & \textbf{22.03} \\
    Variant CVAE          & 16.07 & 6.76 & 1.78 & 15.61 & 3.28 & 20.58 \\
    % Structure (c) & 15.85 & 6.53 & 1.69 & 15.25 & \textbf{3.80} & \textbf{22.67} \\
    % Model                    & PPL $\downarrow$   & B-2 $\uparrow$  & B-4 $\uparrow$  & R-L $\uparrow$  & D-1 $\uparrow$    \\
    % \midrule
    % Structure (b)          & 16.07 & 6.76 & 1.78 & 15.61 & 3.28  \\
    % Structure (c) & 15.85 & 6.53 & 1.69 & 15.25 & \textbf{3.80}  \\
    % PoKE       & \textbf{15.84} & \textbf{6.79} & \textbf{1.78} & \textbf{15.84} & 3.73  \\
    \bottomrule
    \end{tabular}}
\end{table}

\begin{figure}[t] 
  \centering
  \includegraphics[width=\linewidth]{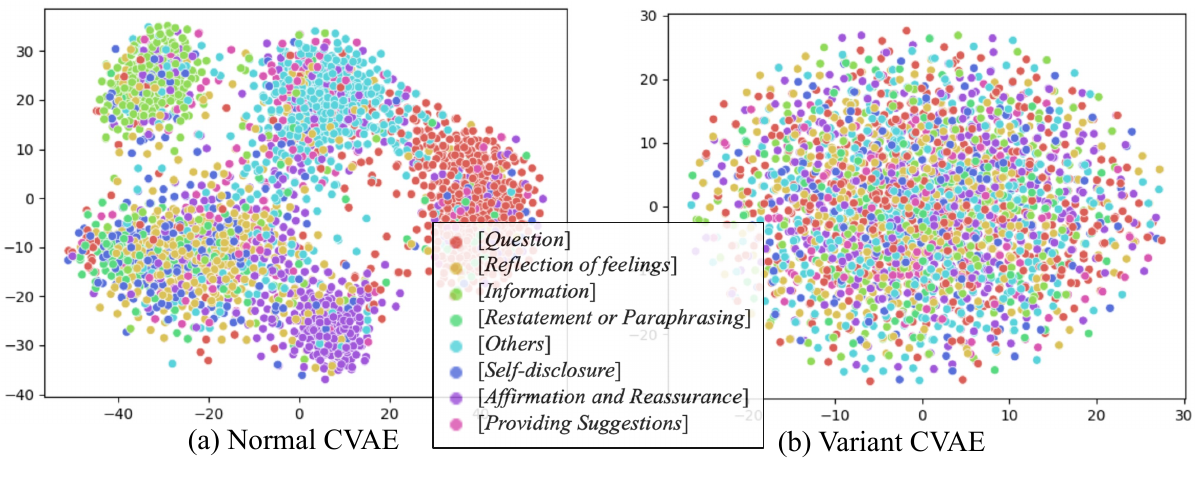}
  \caption{t-SNE visualization of the posterior z for test responses with 8 strategies. (a) Normal CVAE: strategy only acts as output to regularize the latent space. (b) Variant CVAE: strategy is only as input condition of latent variable.} \label{fig:cvae}
\end{figure}

\subsection{Effect of CVAE Structure}
In this section, we adjust the structure of CVAE to explore the reasonable manner of utilizing strategy.
For normal CVAE, namely the PoKE, strategy is only used as the output to regularize the latent space (Eq. (\ref{eq:strategy})). Here, we consider a variant CVAE that strategy is merely as input condition to model latent variable, i.e. the recognition network becomes $q_\phi(\mathbf{z}|x, r, y)$ and $\mathcal{L}_{y}$ is ignored. 
% The difference between graphical models of these two structures are shown in Figure \ref{fig:graph}. 
We conduct quantitative and visualization experiments to compare these two structures of CVAE. 

Table \ref{tb:cvae} shows that the overall performance of variant CVAE drops a lot, especially in diversity. Meanwhile, the visualization in Figure \ref{fig:cvae}(b) exhibits that the latent space is independent of strategy, so strategy information is vanished from the latent variable. This demonstrates that only taking strategy as input is inadequate to model an informative latent space.
%  and latent variable merely depends on other input conditions.
In contrast, PoKE has a better diversity (Table \ref{tb:cvae}) and can learn a meaningful latent space highly correlated with the support strategy (Figure \ref{fig:cvae}(a)). 
This demonstrates that PoKE effectively regularizes the latent space and incorporates the informative latent variable into decoder to generate diverse responses.

\begin{table}[t]
  \caption{The results of ablation study on PoKE variants.} \label{tb:ablation}
  \begin{tabular}{lcccccc}  
    \toprule
    % Model                    & PPL $\downarrow$   & B-2 $\uparrow$  & B-4 $\uparrow$  & R-L $\uparrow$  & D-1 $\uparrow$  & D-2 $\uparrow$  \\
    \textbf{Model}                    & \textbf{PPL} $\downarrow$   & \textbf{B-2} $\uparrow$  & \textbf{B-4} $\uparrow$  & \textbf{R-L} $\uparrow$  & \textbf{D-1} $\uparrow$  & \textbf{D-2} $\uparrow$  \\
    \midrule
    PoKE w/o $\mathbf{e}$       & \textbf{15.74} & 6.59 & 1.63 & 15.54 & 3.53 & 21.45 \\
    PoKE w/o $\mathbf{T}$          & 15.84 & 6.66 & 1.70 & 15.80 & 3.67 & 21.65 \\
    PoKE w/o $\mathbf{z}$ & 16.13 & 6.57 & 1.64 & 15.42 & 3.36 & 20.59 \\
    % PoKE w/o $\mathbf{s}$ &   &   & 1.64 & 15.42 & 3.36 & 20.59 \\
    \midrule
    PoKE                     & 15.84 & \textbf{6.79} & \textbf{1.78} & \textbf{15.84} & \textbf{3.73} & \textbf{22.03} \\
    \bottomrule
    \end{tabular}
\end{table}

\begin{table*}[ht!]
  \caption{An example of ESConv and the responses generated from PoKE and other SOTA models. The retrieved context-related exemplars are also displayed. \textcolor{orange}{Orange text} denotes the strategy label of exemplar. Relevant parts have the same color.} \label{tb:case}
  \centering
  \small  
  \begin{tabular}{l|p{0.85\textwidth}l}
  \toprule
  % \multicolumn{2}{c}{\textbf{Situation}}\\
  \textbf{Situation} & My partner broke up with me, but he did not tell me the reason why. \\
  % \hline
  \midrule
  % \multicolumn{2}{l}{My partner broke up with me, but he did not tell me the reason why.}\\
  % Seeker &\makecell[l]{My boyfriend and I recently broke up due to long-distance relationship and the impact COVID-19...}\\%%has had on the travel industry.}\\
  % \hline
  % \multicolumn{2}{c}{\textbf{Context}}\\
  % \hline
  % \multicolumn{2}{l}{\textit{Seeker}: I feel like I am in depression.}\\
  % \multicolumn{2}{l}{\textit{Supporter}: \textcolor{orange}{[\textit{Question}]} I'm sorry to hear that. What do you think may have led up to these feelings of being depressed?}\\
  % \multicolumn{2}{l}{\textit{Seeker}: My partner broke up with me impromptu.}\\
  % \multicolumn{2}{l}{\textit{Supporter}: \textcolor{orange}{[\textit{Self-disclosure}]} I've been through a few bad breakups myself in the past, so I know how you must be feeling right now.}\\
  % \multicolumn{2}{l}{\textit{Seeker}: How did you deal with them, and specifically what was your coping mechanism?}\\
        & {\textit{Seeker}: I feel like I am in depression.}\\
        & {\textit{Supporter}: \textcolor{orange}{[\textit{Question}]} I'm sorry to hear that. What do you think may have led up to these feelings of being depressed?}\\
  \textbf{Context} & {\textit{Seeker}: My partner broke up with me impromptu.}\\
        & {\textit{Supporter}: \textcolor{orange}{[\textit{Self-disclosure}]} I've been through a few bad breakups myself in the past, so I know how you must be feeling right now.}\\
        & {\textit{Seeker}: How did you deal with them, and specifically what was your coping mechanism?}\\
  % \hline
  \midrule 
  % Seeker &\makecell[l]{I feel like I am in depression.}\\
  % \hline
  % Supporter &\makecell[l]{I'm sorry to hear that. What do you think may have led up to these feelings of being depressed?} \\
  % \hline
  % Seeker & \makecell[l]{My partner broke up with me impromptu.} \\
  % \hline
  % Supporter &\makecell[l]{I've been through a few bad breakups myself in the past, so I know how you must be feeling right now.}\\
  % \hline
  % Seeker & \makecell[l]{How did you deal with them, and specifically what was your coping mechanism?}\\
  % \hline
  % \multicolumn{2}{c}{\textbf{Exemplars}}\\
  % \hline
  % \multicolumn{2}{l}{\textcolor{orange}{[\textit{Affirmation and Reassurance}]} \textcolor{blue}{\textbf{I can understand your situation}}. You think that its normal for you, but other has different outlook.}\\
  % % \multicolumn{2}{p{0.85\textwidth}}{\textcolor{orange}{[\textit{Self-disclosure}]} I know that when I broke up with my partner, I spent at least a few days crying and trying to figure out what my life would be now.}\\
  % % \multicolumn{2}{c}{...}\\
  % \multicolumn{2}{p{0.9\textwidth}}{\textcolor{orange}{[\textit{Providing suggestions}]} 
  % I tried to focus on other areas of my life and to \textcolor{olive}{\textbf{critically reflect on my relationship}}.
  %  Largely, I threw myself into my work and tried to better that aspect of my life, and \textcolor{teal}{\textbf{hung out with people}} that I knew cared about me.}\\
  \textbf{Exemplars} &\textcolor{orange}{[\textit{Affirmation and Reassurance}]} \textcolor{blue}{\textbf{I can understand your situation}}. You think that its normal for you, but other has different outlook.\\
  % \multicolumn{2}{p{0.85\textwidth}}{\textcolor{orange}{[\textit{Self-disclosure}]} I know that when I broke up with my partner, I spent at least a few days crying and trying to figure out what my life would be now.}\\
  % \multicolumn{2}{c}{...}\\
          & ... \\
          & \textcolor{orange}{[\textit{Providing suggestions}]} 
  I tried to focus on other areas of my life and to \textcolor{olive}{\textbf{critically reflect on my relationship}}.
   Largely, I threw myself into my work and tried to better that aspect of my life, and \textcolor{teal}{\textbf{hung out with people}} that I knew cared about me.\\
  % \multicolumn{2}{l}{better that aspect of my life, and hung out with people that I knew cared about me.}\\
  % \multicolumn{2}{l}{\textcolor{orange}{[\textit{Providing suggestions}]}
  % [Self-disclosure] I think this is to be expected from a very fresh break-up. I know that when I broke up with my partner, I spent at least a few days crying and trying to figure out what my life would be now. 
  % \hline
  \midrule
  % \multicolumn{2}{c}{\textbf{Response}}\\ \hline
  BlenderBot-Joint & \makecell[l]{I have been able to cope with my ex by talking to him about my feelings. I have found that talking to people helps me to cope.}\\ 
  % \hline
  \midrule
  MISC$^{*}$  &\makecell[l]{Have you considered seeing a therapist?}\\
  % \hline
  \midrule
  PoKE  &\makecell[lp{0.85\textwidth}]{\textcolor{blue}{\textbf{I can definitely relate to that}}. I have been in that situation, and it really helped to \textcolor{teal}{\textbf{get outside and have a good conversation}} about why I feel that way. I think that it is important to remember to \textcolor{olive}{\textbf{think about the bright side of your relationship}}, too.}\\
  % \hline
  \midrule
  \textbf{\emph{Gold Response}} & \makecell[lp{0.85\textwidth}]{Ha, well at first I cried and was mad and just moped and hated everything, but eventually I got tired of no getting anything out of my days. I slowly realized that I had put myself in slow motion and then slowly started to do things I used to do to maybe come out of it.}\\
  \bottomrule
  \end{tabular}
  \label{tab:generate example}       % Give a unique label
\end{table*}

\subsection{Ablation Study}
\label{sec:ablation}

To understand the importance of prior knowledge and latent variable for providing better emotional support, we conduct an ablation study to investigate the effect of the key components in PoKE.
% In this section, we perform ablation study on ESConv to investigate the effect of the key components in PoKE. 
We design several variants of PoKE by removing some specific parts:

\noindent \textbf{PoKE w/o $\mathbf{e}$.} Remove the prior knowledge of exemplars, i.e. the the denoised exemplars vector $\mathbf{e}$ is excluded from \textit{memory} vectors.

\noindent \textbf{PoKE w/o $\mathbf{T}$.} Remove the prior knowledge of strategy sequence, i.e. the first-order Markov transition matrix $\mathbf{T}$ of strategy is ignored when modeling the distribution of strategy.

% Only use latent variable to model the distribution of strategy.
\noindent \textbf{PoKE w/o $\mathbf{z}$.} The CVAE module is removed, and we directly use input conditions instead of latent variable to predict the strategy. In addition, the latent variable $\mathbf{z}$ is removed from \textit{memory} vectors. 
% \item \textbf{PoKE w/o $\mathbf{s}$.} The representation of the mixed strategy $\mathbf{s}$ is excluded from \textit{memory} vectors.
%  1) prior knowledge of exemplar (called \textbf{PoKE w/o Exemplars}), 2) prior knowledge of strategy sequence (called \textbf{PoKE w/o Markov}), 3)

Table \ref{tb:ablation} shows the results of ablation studies. We can find that almost all variants perform worse than the PoKE, which verifies each component in PoKE. 
The results of PoKE w/o $\mathbf{e}$ and w/o $\mathbf{T}$ show that both generation quality and diversity get worse after removing prior knowledge. 
This suggests that explicitly using prior knowledge in historical conversations benefits more relevant responses, and plenty of various exemplars help generate responses with higher diversity.
However, compared to PoKE, the PPL of PoKE w/o $\mathbf{e}$ improves slightly. We speculate that exemplars contain some token-level noise, thus impairing fluency. We leave the research of denoising exemplars at the token-level as future work. 
Regarding the PoKE w/o $\mathbf{e}$, D-1 and D-2 drop by a large margin. This result is as expected because the latent variable models the one-to-many mapping relationship of strategy. By sampling latent variable, randomness is introduced to strategy distribution and enables infrequent strategies to be considered. 
% Moreover, the usage of mixed strategy and memory schema incorporates these various strategies into the decoder to generate a more diverse response.

\section{Case Study}

Table \ref{tb:case} shows an example of ESConv and the responses generated from PoKE and other SOTA models. 
From the seeker's situation, we can know the seeker has emotional stress of breaking up with his partner, and he is asking for suggestions. 
BlenderBot-Joint directly provide a suggestion, but it is not suitable or commonly used. 
MISC uses the COMET to infer the commonsense that seeing a therapist may help overcome the problem and utilizes it for guiding generation, but it does not combine its own experience. 
The gold reference shares his solutions of getting rid of emotional stress.
Compared with them, PoKE makes a better response thanks to latent variable and prior knowledge. 
% Benefiting from modeling strategy distribution by latent variable, 
PoKE expresses a mixed strategy smoothly, i.e. affirming the seeker before sharing advice. Additionally, PoKE utilizes abundant reference information about strategy expression and suggestions from exemplars explicitly or implicitly. For instance, (1) ``I can definitely..." expresses the strategy of \textit{Affirmation and Reassurance} by explicitly referring to the sentence pattern of the first exemplar, and (2) ``get outside ..." as well as ``think about ..." implicitly incorporate the suggestions of the last exemplar into the response. 
Besides, we visualize the correlation between the prior knowledge of strategy and the predicted strategy distribution in Figure \ref{fig:distribution}, which is detailed in Appendix \ref{app:transition}.  
%  for the above example in Figure \ref{fig:distribution} (Appendix \ref{app:transition})to show that the prior knowledge of strategy transition can serve as a basis to help make a reasonable decision about strategy.

% Retrieved exemplary responses give supporter more clues to focus on seeker's problem and express strategy more accurately.

\section{Conclusion}

In this paper, we explore the emotional support conversation under the setting of no external knowledge and propose PoKE, a prior knowledge enhanced model with latent variable to provide emotional support in conversation. 
The proposed PoKE could utilize the prior knowledge in terms of exemplars and strategy sequence, and models the one-to-many mapping relationship of strategy. 
Then, PoKE utilizes strategy distribution to denoise the exemplars and applies a memory schema to incorporate encoded information into decoder.
The experiments on automatic and human evaluation demonstrate the superiority and diversity of PoKE without external knowledge. Moreover, the analytical experiments prove that PoKE can effectively utilize prior knowledge to generate better emotional support and learn an informative latent variable to respond with high diversity. 
In future work, we will further refine our model to outperform the methods using external knowledge and explore the manner of efficiently incorporating external knowledge.
% In future work, we will further explore the exemplars denoised at the token-level to refine the generation.
% take a deep look at  leave the research of denoising exemplars at token-level as a future work. 

% abundant prior knowledge is conducive to high-quality emotional support,
% and rely on CVAE to emphasize on strategy-related exemplars to refine expression of strategy. 

% and siginificantly outperforms several existing state-of-the-art methods for the inductive link prediction task, 

% PoKE utilizes prior knowledge to facilitate exploring user's problem and making decision of strategy, and relies on CVAE to pay more attention on strategy-related exemplars at sequence-level to improve expression of strategy. 

\clearpage

\bibliographystyle{ACM-Reference-Format}
% \balance
\bibliography{sample-base}

\appendix

\section{ESConv Dataset}
The detailed statistics of the original ESConv are shown in Table \ref{tb:esconv}.
The long average length of turns (29.8) indicates that the ESC task needs more turns to provide an effective emotional support for seeker.

\begin{table}[htbp]
  \caption{Statistics of ESConv.} 
  \label{tb:esconv}
  \begin{tabular}{lccc}
    \toprule
    Category & Total & Support & Seeker \\
    \midrule
    \# Dialogues  & 1,053  & - & - \\
    \# Utterances & 31,410 & 14,855  & 16,555 \\

    % \# Situations
    Avg. length of turns & 29.8 & 14.1 & 15.7 \\
    Avg. length of utterances & 17.8  & 20.02 & 15.7 \\
    Avg. length of situations &  22.85 & - & - \\
    \bottomrule
  \end{tabular}
\end{table}

% \subsection{Definitions of Support Strategies} 
% We list the definitions of 8 support strategies in ESConv with reference to  \cite{DBLP:conf/acl/LiuZDSLYJH20}.

% \noindent \textit{\textbf{Question.}} Asking for information related to the problem to help the help-seeker articulate the issues that they face. Open-ended questions are best, and closed questions can be used to get specific information.

% \noindent \textit{\textbf{Restatement or Paraphrasing.}} A simple, more concise rephrasing of the help-seeker's statements that could help them see their situation more clearly.

% \noindent \textit{\textbf{Reflection of Feelings.}} Articulate and describe the help-seeker's feelings.

% \noindent \textit{\textbf{Self-disclosure.}} Divulge similar experiences that you have had or emotions that you share with the help-seeker to express your empathy.

% \noindent \textit{\textbf{Affirmation and Reassurance.}} Affirm the help-seeker's strengths, motivation, and capabilities and provide reassurance and encouragement.

% \noindent \textit{\textbf{Providing Suggestions.}} Provide suggestions about how to change, but be careful to not overstep and tell them what to do.

% \noindent \textit{\textbf{Information.}} Provide useful information to the help-seeker, for example with data, facts, opinions, resources, or by answering questions.

% \noindent \textit{\textbf{Others.}} Exchange pleasantries and use other support strategies that do not fall into the above categories.

\section{Implementation Details} \label{app:implementation}
Similar to BlenderBot-Joint  \cite{DBLP:conf/acl/LiuZDSLYJH20} and MISC \cite{DBLP:conf/acl/TuLC0W022}, we use BlenderBot Small  \cite{roller2021recipes} as our model's backbone. 
The default size of hidden state $d_h$ in BlenderBot Small is 512, and the dimension of latent variable $d_z$ is set as 64 by parameter search. 
According to the result in Section \ref{sec:exem}, we retrieve $k = 10$ exemplars for each context. 
The coefficient $\lambda$ in Eq. (\ref{eq:objective}) is set to 1.0. 
For stable optimization, the total KL annealing steps with 10000, strategy annealing rate $\beta$ with $1 \times 10^{-3}$ and steps $T$ with 1000 achieves the best performance. 
The batch size of training and validation is set to 20 and 50 respectively. 
We use optimizer AdamW  \cite{DBLP:journals/corr/abs-1711-05101} to optimize our model.
We train the model for 8 epochs and select the best models based on the perplexity of the validation data.
For decoding, we employ Top-$k$ and Top-$p$ sampling methods in previous work  \cite{DBLP:conf/acl/LiuZDSLYJH20}, and set $k=30$, $p=0.9$, temperature $\tau = 0.9$ and repetition penalty to 1.03.
For a fair comparison, all methods are implemented using the same hyperparameters and on the Tesla V100 GPU.
% Regarding latent variable, we set its dimension $d_z$ as 

%  to help model current strategy distribution.
 
\begin{figure}[htbp] 
  \centering
  \includegraphics[width=0.95\linewidth]{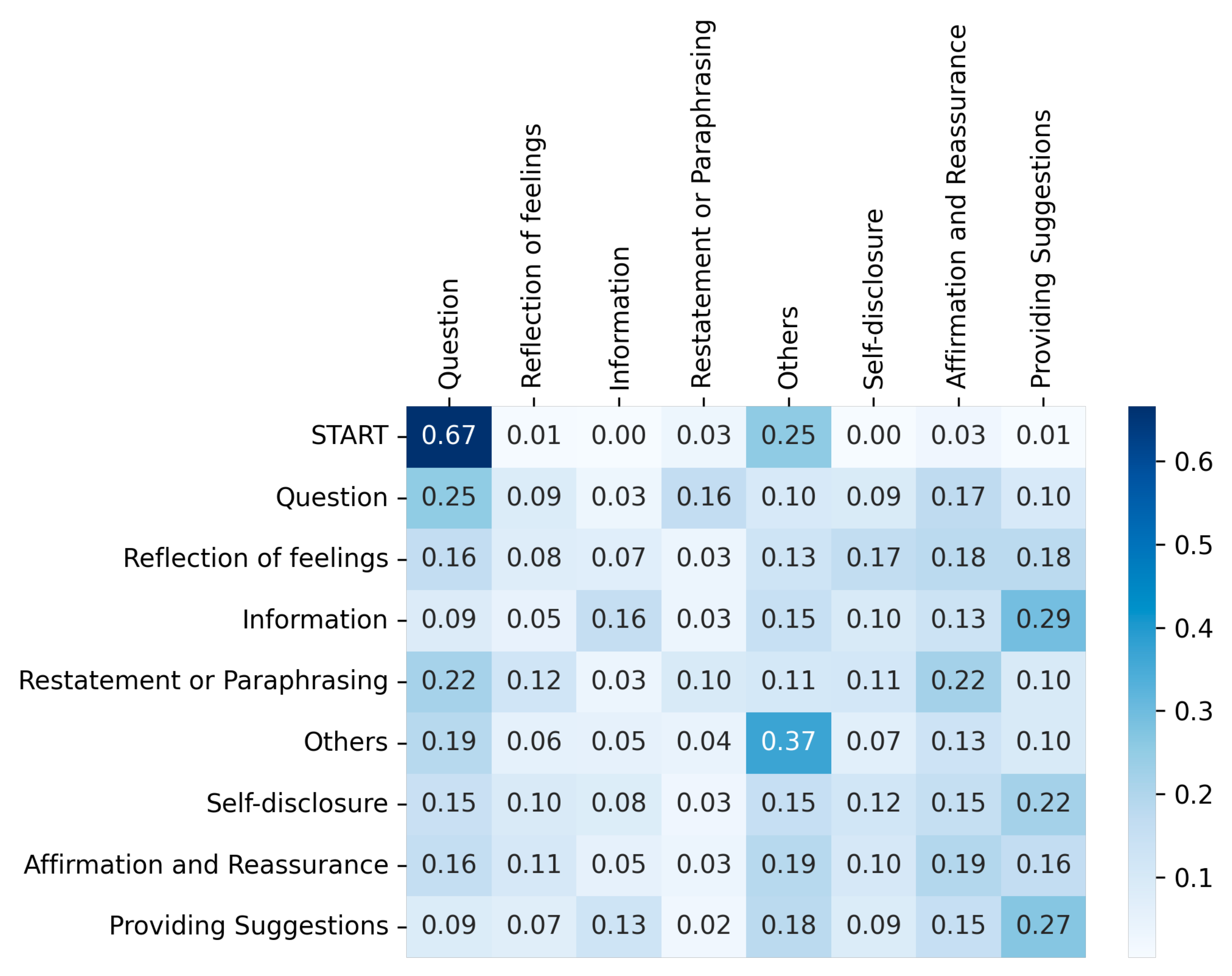}
  \caption{First-order Markov transition matrix $\mathbf{T}$ of strategy calculated in training set. START means the current conversation turn is the first round, and there is no previous strategy.} \label{fig:markov}
\end{figure}

\section{Prior Knowledge of Strategy}
\subsection{Markov Transition Matrix of Strategy} \label{app:markov}
The first-order Markov transition matrix $\mathbf{T} \in \mathbb{R}^{(m+1) \times m}$ of strategy calculated in the training set is shown in Figure \ref{fig:markov}.
The transition matrix $\mathbf{T}$ containing prior knowledge of strategy selection is simple but practical in ESC task, which is demonstrated in Section \ref{sec:ablation}.
From this matrix, we can find useful prior knowledge about general patterns of strategy selection. 
For instance, supporters tend to take \textit{Question} as a conversation starter to acquire more seeker's information. 
After sharing the similar difficulties they faced, supporters tend to use \textit{Providing suggestions} to give advice based on their experience, and so on.

\subsection{Applied in Case Study}
\label{app:transition}
\begin{figure}[t] 
  \centering
  \includegraphics[width=\linewidth]{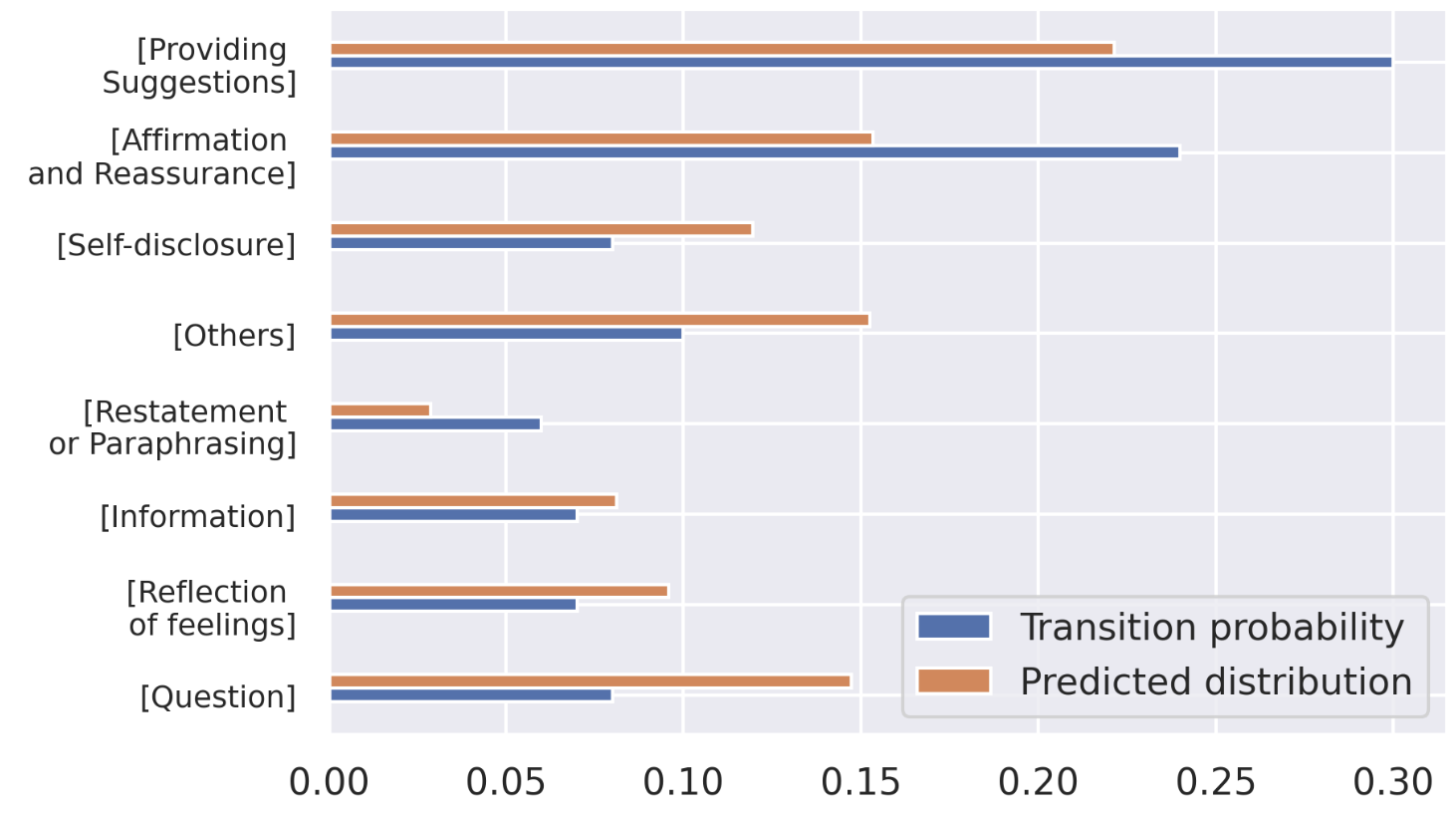}
  \caption{The visualization of transition probability of the previous strategy \textit{Self-disclosure} taken by the supporter and the predicted distribution in case study.} \label{fig:distribution} 
\end{figure}

For the case in Table \ref{tb:case}, we visualize the correlation between the prior knowledge of strategy and the predicted strategy distribution in Figure \ref{fig:distribution}.
In that case, the previous strategy taken by the supporter is \textit{Self-disclosure}. According to the  first-order Markov transition matrix $\mathbf{T}$ in Figure \ref{fig:markov}, we can obtain the transition probability of the strategy \textit{Self-disclosure}. Besides, we use Eq. (\ref{eq:distribution}) to predict the strategy distribution via latent variable and transition probability. 
Figure \ref{fig:distribution} shows that the two distributions have a similar pattern, such as the maximum probability of \textit{Providing Suggestions} and the most unlikely strategy \textit{Restatement or Paraphrasing}. This indicates that the simple transition matrix of strategy can provide practical prior knowledge for current strategy decisions. Moreover, according to the predicted strategy distribution, PoKE can further adjust strategy distribution based on the current context (e.g. higher probability of \textit{Question} and \textit{Self-disclosure}). 

% for the above example in Figure \ref{fig:distribution} (Appendix \ref{app:transition})to show that the prior knowledge of strategy transition can serve as a basis to help make a reasonable decision about strategy.

\section{Conditional Variable Autoencoder}
\label{app:cvae}

Mathematically, our goal is to maximize the conditional likelihood of response $r$ for the given conditions $x$: 
\begin{equation}
  p(r|x) = \int p(r| \mathbf{z}, x)p(\mathbf{z}|x) d\mathbf{z}, 
\end{equation}
where $p(\mathbf{z}|x)$ involves an intractable marginalization over the latent variable $\mathbf{z}$. To solve that probelm and model the latent variable, CVAE uses a prior network $p_\theta(\mathbf{z}|x)$ to approximate $p(\mathbf{z}|x)$, and a recognition network $q_\phi(\mathbf{z}|x, r)$ to approximate true posterior $p(\mathbf{z}|x, r)$. 
In general, the latent variables from prior network and recognition network are assumed to fit multivariate Gaussian distribution with a diagonal covariance matrix, i.e. $p_\theta(\mathbf{z} | x) \sim \mathcal{N}(\boldsymbol{\mu}, \boldsymbol{\sigma}^2 \mathbf{I})$ and $q_\phi(\mathbf{z} |r,x) \sim \mathcal{N}(\boldsymbol{\mu}^\prime, \boldsymbol{\sigma}^{\prime2} \mathbf{I})$.
Then, CVAE can be trained by maximizing a variational lower bound, consisting of two terms: negative likelihood loss of decoder and K-L regularization: 
\begin{align}
  \mathcal{L}_{ELBO}(\theta, \phi ; r, x) &= \mathcal{L}_{nll} + \mathcal{L}_{kl} \nonumber \\
  &=\mathbf{E}_{q_\phi(\mathbf{z} | x, r)}\left[\log p_\theta(r | \mathbf{z}, x)\right]  \nonumber \\
  &-K L\left(q_\phi(\mathbf{z} | r, x) \| p_\theta(\mathbf{z} | x)\right) \nonumber \\
  & \leq \log p(r | x),
\end{align}
where $p_\theta(r | \mathbf{z}, x)$ is the decoder network for generation, which is illustrated in Section \ref{sec:decoder}. 

In CVAE, both the prior network and recognition network apply the structure of multilayer perceptron, and then we can calculate the mean $\boldsymbol{\mu} \in \mathbb{R}^{d_z}$ and variance $\boldsymbol{\sigma} \in \mathbb{R}^{d_z}$ in multivariate Gaussian distribution by:
\begin{align}
  \left[\begin{array}{c}
    \boldsymbol{\mu} \\
    \log \left(\boldsymbol{\sigma}^{2}\right)
  \end{array}\right]&=\operatorname{MLP}_p(x) = \mathbf{W}_p[\mathbf{c} ; \mathbf{s} ; \mathbf{p}] + \mathbf{b}_p, \\
  \left[\begin{array}{c}
  \boldsymbol{\mu}^{\prime} \\
  \log \left(\boldsymbol{\sigma}^{\prime 2}\right)
  \end{array}\right]&=\operatorname{MLP}_q(x, r) = \mathbf{W}_q[\mathbf{c} ; \mathbf{s} ; \mathbf{p} ; \mathbf{r}] + \mathbf{b}_q,
\end{align}
where $\mathbf{W}_p \in \mathbb{R}^{2d_z \times 3d_h}, \mathbf{b}_p \in \mathbb{R}^{2d_z}, \mathbf{W}_q \in \mathbb{R}^{2d_z \times 4d_h}, \mathbf{b}_q \in \mathbb{R}^{2d_z}$, and $\mathbf{r}$ is the representation of response reference obtained in the similar way to Eq. (\ref{eq:H}) and Eq. (\ref{eq:h}).
Then we use the reparameterization trick  \cite{DBLP:journals/corr/KingmaW13} to sample latent variable $\mathbf{z}$. During training, we sample latent variables from the recognition network and prior network to optimize the CVAE by Eq. (\ref{eq:cvae}). While during inference, there is no response reference, so we only sample latent variable from the prior network and pass it to the decoder for generation. For more mathematical details, please refer to  \cite{DBLP:journals/corr/KingmaW13}.

\end{document}